\documentclass[twocolumn,switch,times]{article}
\usepackage{preprint}
\usepackage{hyperref}
\usepackage[authoryear,round]{natbib}

\hypersetup{
    colorlinks=true,
    linkcolor=black,
    urlcolor=cyan,
    citecolor=cyan,
    anchorcolor=black
}

\usepackage[utf8]{inputenc}	
\usepackage[T1]{fontenc}
\usepackage{xcolor}
\usepackage{lineno}					
\usepackage{tikz} 					
\usepackage{fancyhdr}
\usepackage{geometry}
\usepackage{lipsum}
\usepackage{scalerel}
\usepackage{sidecap}
\usepackage{times}
\usepackage{doi}
\usepackage{amssymb}
\usepackage{amsmath}
\usepackage{tabularx}
\usepackage{dsfont}
\usepackage{bbm}
\usepackage{bm}
\usepackage{array,booktabs}
\usepackage{upgreek}
\usepackage{algorithm}
\usepackage{algpseudocodex}
\usepackage[title]{appendix}
\usepackage{authblk}
\usepackage{placeins}
\usepackage{cprotect}

\usepackage{newfloat}
\DeclareFloatingEnvironment[name={Supplementary Figure}]{suppfigure}
\usepackage{sidecap}
\sidecaptionvpos{figure}{c}
\usepackage[font=small]{caption}
\usepackage{etoolbox}

\usepackage{titlesec}
\titlespacing\section{0pt}{12pt plus 3pt minus 3pt}{1pt plus 1pt minus 1pt}
\titlespacing\subsection{0pt}{10pt plus 3pt minus 3pt}{1pt plus 1pt minus 1pt}
\titlespacing\subsubsection{0pt}{8pt plus 3pt minus 3pt}{1pt plus 1pt minus 1pt}

\definecolor{lime}{HTML}{A6CE39}

\newcolumntype{P}[1]{>{\centering\arraybackslash}p{#1}}

\newcommand{\appendixref}[1]{%
  \hyperref[#1]{Appendix~\ref*{#1}}\unskip
}

\title{Multispectral airborne laser scanning dataset for tree species classification: MS-ALS-SPECIES}

\usepackage{authblk}

\author[1,\small\textdagger]{Matti Hyyppä}

\author[1,\small\textdagger,\small\ensuremath{\ast}]{Klaara Salolahti}

\author[1]{Eric Hyyppä}

\author[1]{Xiaowei Yu}

\author[1]{Josef Taher}

\author[1]{Leena Matikainen}

\author[1]{Matti Lehtomäki}

\author[1]{Paula Litkey}

\author[1]{Teemu Hakala}

\author[1]{Harri Kaartinen}

\author[1]{Juha Hyyppä}

\author[1]{Antero Kukko}

\affil[1]{Department of Remote Sensing and Photogrammetry, Finnish Geospatial Research Institute FGI, The National Land Survey of Finland, Vuorimiehentie 5, Espoo, FI-02150, Finland}

\begin{document}

\twocolumn[\begin{@twocolumnfalse}

\maketitle

\begin{abstract}
The shift from stand-level to individual-tree-level forest assessments supports improved species mapping and biodiversity monitoring, particularly in boreal ecosystems where tree species like aspen (\textit{Populus tremula} L.) play a keystone role. Airborne laser scanning (ALS) is the standard for such inventories, but a major limitation for developing improved species classification methods is the small number of publicly available ALS datasets containing high-quality, field-validated reference data. Recently, multispectral ALS data has shown promise for tree species classification, but the progress is hindered by the lack of open multispectral ALS datasets with high-quality field reference data. This paper presents and details an open multispectral ALS dataset for tree species classification that was used before its public release for an international benchmarking study of machine learning and deep learning classification methods in a related publication by \citet{taher2025}. The dataset comprises 6326 segment-level point clouds of individual trees representing nine species in southern Finland. The point cloud data has been acquired using two multispectral laser scanning systems each operating at three laser wavelengths: a helicopter-borne system (HeliALS) with a point density exceeding 1000~$\mathrm{points}/\mathrm{m}^2$ and an Optech Titan system with approximately 35~$\mathrm{points}/\mathrm{m}^2$. Furthermore, we present a crowdsourcing application that facilitates the collection of high-quality field reference data of tree species in an efficient and scalable manner.
Our article showcases the versatility of the open dataset by  presenting new analyses on species classification using multispectral data building upon the initial findings of \citet{taher2025}. We offer an improved benchmark of classification methods by evaluating the classification accuracy exclusively on tree segments with minimal segmentation errors, which is shown to
improve the classification accuracy by 2--3 percentage points.
The ranking of the classification methods remains largely unchanged and the point transformer model still achieves the highest accuracy, thus confirming the results of the prior benchmarking study.     
Furthermore, we study the relation between classification accuracy and tree height to demonstrate the advantage of the point transformer model for small trees and minority species.
The public release of this comprehensive dataset and the related methodology facilitate the development of deep learning models for accurate tree species classification using multispectral laser scanning, while also advancing research in ecological monitoring and sustainable forest management.
\end{abstract}

\keywords{Lidar, Tree species, Classification, Multispectral Laser Scanning, Open Dataset, Segmentation}

\vspace{6.0cm}

\vfill\hrule\vspace{2pt}
\enlargethispage{2\baselineskip}
\footnotesize Published in ISPRS Open Journal of Photogrammetry and Remote Sensing, 2026. Final version available at: \url{https://doi.org/10.1016/j.ophoto.2026.100142} .
\copyright \ 2026. This manuscript version is made available under the CC-BY-NC-ND 4.0 license \url{https://creativecommons.org/licenses/by-nc-nd/4.0/} .
MS-ALS-SPECIES dataset is publicly available in Zenodo: \url{https://doi.org/10.5281/zenodo.17077255}\\

{\large\textsuperscript{\ensuremath{\ast}}}Corresponding author\\
{\large\textsuperscript{\textdagger}}These authors contributed equally.\\
\textit{Email addresses}: matti.hyyppa@nls.fi (Matti Hyyppä), klaara.salolahti@nls.fi (Klaara Salolahti), eric.hyyppa@nls.fi (Eric Hyyppä), xiaowei.yu@nls.fi (Xiaowei Yu), josef.taher@nls.fi (Josef Taher), leena.matikainen@nls.fi (Leena Matikainen), matti.lehtomaki@nls.fi (Matti Lehtomäki), paula.litkey@nls.fi (Paula Litkey), teemu.hakala@nls.fi (Teemu Hakala), harri.kaartinen@nls.fi (Harri Kaartinen), juha.hyyppa@nls.fi (Juha Hyyppä), antero.kukko@nls.fi (Antero Kukko)

\end{@twocolumnfalse}]


\section{Introduction}
\label{sec:introduction}

Knowledge of tree species is important for effective management of forests for multiple purposes, from harvesting to conservation and even finding wild berries \citep{turtiainen2015}. The variety of tree species is a key factor in shaping forest ecosystems, impacting their productivity, resilience, recovery, competition, health, economic potential, and biodiversity \citep{jonsson2019levels, baeten2019identifying}. For example,  aspen (\textit{Populus tremula} L.) is a keystone species in the boreal forest zone, because it has a vital role in supporting biodiversity by hosting a wide variety of fungi, lichens, insects, birds, and mammals \citep{kuusinen1994epiphytic, angelstam1994woodpecker,tikkanen2006red, remm2017multilevel}.

To date,  forest inventories based on airborne laser scanning (ALS) have predominantly focused on the area-based approach 
in boreal ecosystems. However, there is a growing interest to acquire inventory data also at the individual tree level  \citep{hyyppa1999method,hyyppa1999detecting,hyyppa2024concepts, soininen2025transferability}. 
This shift is driven, in part, by the need for more detailed biodiversity mapping. Rare broad-leaved species \citep{toivonen2024mapping}, old-growth trees, and both standing \citep{kaminska2018species, amiri2019classification} and downed coarse woody debris \citep{yrttimaa2019detecting} serve as critical indicators of forest biodiversity, necessitating a finer-scale inventory approach at the individual tree level. 

Recently, we performed an international benchmarking study on tree species classification using two multispectral ALS datasets (35 $\mathrm{pts}/\mathrm{m}^2$ and 1300 $\mathrm{pts}/\mathrm{m}^2$) \citep{taher2025}. The primary aims of the study were to compare various state-of-the-art deep learning and machine learning classifiers, and to investigate the impact of multi-channel spectral features on classification accuracy. In line with previous studies \citep{orka2009classifying, axelsson2018exploring, kukkonen2019multispectral, hakula2023individual}, our study demonstrated a significant improvement in individual-tree-level species classification accuracy by utilizing single-channel and multi-channel intensity data in addition to the point cloud geometry. On the denser ALS data, the addition of single-channel spectral information increased the overall classification accuracy for nine species from 73.0\% to 85.9\% and the macro-average accuracy from 48.9\% to 65.1\% when using a point transformer neural network \citep{zhao2021point}. The overall accuracy increased further to 87.9\% and the macro-average accuracy to 71.2\% when incorporating additional spectral features on two more wavelengths, resulting in three-channel multispectral information.   
By subsampling the point density of the multispectral data, we further
found that multi-channel spectral features were most beneficial at lower point densities on the order of 10 $\mathrm{pts}/\mathrm{m}^2$.

Originally, multispectral lidar was introduced in \citet{kaasalainen2007toward} to enhance object detection and classification from ALS point clouds. Since then, laser scanning with several wavelengths has also shown promise for tree species classification \citep{axelsson2018exploring, kukkonen2019multispectral, taher2025}. The advantage of multispectral data arises from the wavelength-dependent spectral response that is distinct for each tree species \citep{hovi2017spectral}. Despite this potential, multispectral ALS data has been scarcely  utilized in studies on deep learning for tree species classification~\citep{wang2024individual, taher2025}, even though recent advancements in deep learning offer promising avenues for semantic point cloud segmentation \citep{reichler2024semantic} and tree species classification \citep{xi2020see, liu2021tree, hell2022classification}. A major obstacle for the development of deep learning classification algorithms for multispectral lidar data has been the lack of publicly available datasets that also include high-quality reference data on tree species.

Two open multispectral ALS datasets have been released in the past. The first dataset was released by Teledyne Optech in 2017 in relation to ISPRS WG III/5. The dataset captured a natural coastal environment using the three-wavelength Optech Titan system \citep{scaioni2018methods}. The second dataset was published in the IEEE GRSS competition in 2018 and included several land cover/use categories \citep{ieee_grs_2018_data_science_comp}. However, no open multispectral ALS datasets currently exist for tree species classification, even though species classification is a key potential use case for multispectral ALS. To date, a handful of single-channel lidar datasets have been publicly released for tree species classification \citep[e.g.,][]{graves_2020_dataset, weiser2022dataset, puliti2025benchmarking, tockner2025tree, korkeala_2025_17639338}. The FOR-species20K dataset by \citet{puliti2025benchmarking} is the largest such dataset to date, comprising terrestrial, mobile, and drone-based laser scanning data. The dataset does not, however, include intensity or reflectance data, because it is mainly aimed at developing automated, sensor-agnostic field reference data collection methods.

 Operational-scale tree species classification is expected to require massive and accurate training datasets because the classification accuracy of deep learning methods scales with training set size following a power law \citep{taher2025}, in analogy to the scaling laws observed in large language models by \cite{kaplan2020scaling}. However, established field inventory methods present a severe bottleneck: they are labor-intensive, time-consuming, and difficult to scale. These traditional approaches primarily rely on five approaches: 1) positioning circular plot centers using a real-time kinematic global navigation satellite system (RTK-GNSS) and using the azimuth angle and distance from the plot center to position individual trees within the plots \citep{mayra2021tree}; 2) using RTK-GNSS to directly position individual trees \citep{mayra2021tree}; 3) utilizing tree maps derived from terrestrial or mobile laser scanning and GNSS measurements \citep{hakula2023individual}; 4) setting up several radio base stations throughout the plot and using trilateration for individual tree positioning, which is the approach the Finnish Forest Centre has integrated into operational forestry \citep{palonen2024puukarttakoealan}; and 5) obtaining tree maps by manual annotation of point clouds \citep{ruoppa2026benchmarking}. While these methods can be used not only for species annotation but also for measuring additional tree attributes, including tree position, tree height, and DBH, they are labor-intensive and require specialized hardware, making them difficult to scale when the primary objective is species classification. Therefore, novel tools are needed to enable the efficient, accurate and scalable collection of tree species reference data in the field. Recently, the topic was discussed by \citet{milenkovic2026tree}, who presented a citizen science app for collecting single-tree information, including tree species.

This paper details an open multispectral ALS dataset for tree species classification that we have made  publicly available via Zenodo in connection to this article, see \citet{hyyppa_2025_zenodo}. The open dataset comprises segment-level point clouds and field-collected species labels of 6326 trees covering nine species scanned with two different multispectral ALS systems in a peri-urban area in the boreal forest zone in southern Finland. The scanner systems include a helicopter-borne HeliALS system with three laser channels yielding a point density of over 1000~$\mathrm{pts}/\mathrm{m}^2$ and  an Optech Titan system with three laser channels mounted on a fixed-wing aircraft providing a point density of approximately 35~$\mathrm{pts}/\mathrm{m}^2$. Before its public release, the dataset was previously used in a related study \citep{taher2025} comparing the classification performance of various state-of-the-art machine learning and deep learning methods. 
While the previous study emphasized classification methods and their performance, the present article focuses on new methodology for scalable field reference data collection, the documentation of the open dataset, additional analyses of factors affecting the classification accuracy, and a comparison of the achieved classification accuracy compared to previous literature.

Overall, the key contributions of this paper include: 
    \begin{enumerate}
        \item Public release of a multispectral ALS dataset for tree species classification: We have made the benchmarking dataset used in \citet{taher2025} publicly available via Zenodo \citep{hyyppa_2025_zenodo} to facilitate future research on tree species classification using deep learning methods. We provide a detailed documentation of the point cloud data and the ground-truth reference data in Sec.~\ref{sec: open dataset}, focusing on important aspects for the user of the dataset.
        Importantly, we present two versions of the field reference data: one matching the setting of the prior benchmarking study and another representing a cleaned dataset after excluding 88 non-ideal tree segments and 18 originally misclassified trees.
        \item Detailed description of novel, scalable field reference collection techniques: We introduce and detail the use of a crowdsourcing application for efficient and scalable collection of high-quality ground-truth data in Secs.~\ref{sec:crowdsourcing application} and ~\ref{sec:field_data_collection}. While there exist open-source tools, such as QField \citep{qfield_software}, for general-purpose field data collection, we chose to develop a dedicated tool specifically tailored for multi-user tree species annotation to improve the efficiency of data collection and for ease of use for domain experts without specialized GIS training.
        \item Detailed analysis of prior art in tree species classification: In Sec.~\ref{sec: related work}, we summarize the state of the art in tree species classification based on single-channel and multispectral ALS to set the results achieved on the published dataset into broader context.
        \item To provide a clearer benchmark for future research with improved automatic segmentation methods, we benchmark the species classification methods studied in \citet{taher2025} on a subset of the test set that includes only clean tree segments  with near ideal segmentation and minimal amount of other vegetation.   Furthermore, we compare the classification accuracy of the best-performing methods as a function of tree height and discuss reasons for typical miclassifications.
    \end{enumerate}

\section{Related work in species classification}
\label{sec: related work}

In this section, we summarize the relevant previous literature on airborne laser scanning for tree species classification, with the focus on multispectral ALS studies and open datasets for tree species classification. A more extensive summary of the past literature on tree species classification using ALS is provided in several review articles, including  \citet{chen2024tree}, \citet{michalowska2021review}, \citet{wang2018review}, \citet{koenig2016full}, and \citet{fassnacht2016review}.

\begin{figure*}[t]
    \centering
    \includegraphics[width=0.85\textwidth]{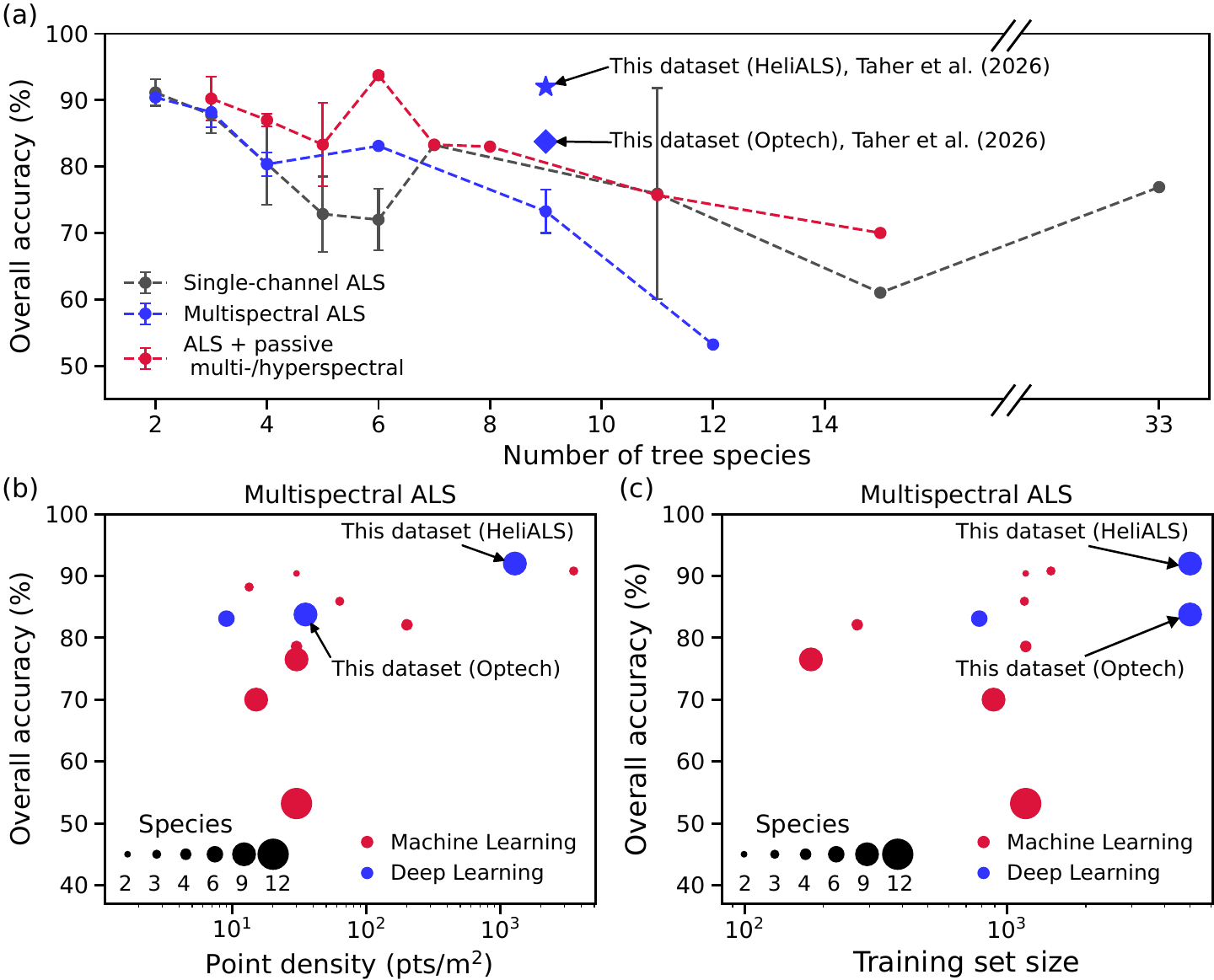}
    \caption{Comparison of the classification accuracy achieved on the published dataset with that reported in previous studies. (a) Median overall accuracy of tree species classification as a function of the number of species for previous studies using single-channel ALS (gray), multispectral ALS (blue), or ALS combined with passive multispectral or hyperspectral imaging (red). The error bars show the median absolute deviation of the accuracy in past studies. The classification accuracy reached in \citet{taher2025} on the HeliALS and Optech Titan datasets published with this paper are shown with a blue star and a blue diamond, respectively. See Appendix~\ref{ap: list of previous studies} for a full list of studies included in the analysis. (b) Overall classification accuracy as a function of point density in the past studies using multispectral ALS. We show the results based on machine learning classifiers in red and the results based on deep learning methods in blue. The marker size is proportional to the number of species considered in the study. (c) Overall classification accuracy as a function of training set size in the past studies using multispectral ALS.     }
    \label{fig: previous literature}
\end{figure*}

In general, the accuracy of tree species classification depends on many factors, including forest-related factors (e.g., number and similarity of tree species, forest maturity, leaf-on/leaf-off conditions), data-related factors (e.g., availability of intensity or spectral information, point density, training set size), and factors related to the chosen classification method (e.g., deep learning vs. machine learning, selected features, data augmentation).  Early studies on ALS-based species classification demonstrated  overall classification accuracies of up to 85--95\% for 2--3 species in the boreal forest zone. These studies applied conventional shallow machine learning classifiers, such as discriminant analysis, in combination with hand-crafted geometric and single-channel intensity features \citep{holmgren2004identifying, holmgren2008species, orka2009classifying, orka2010effects, korpela2010tree}. These early studies focused on species such as pine and spruce, whose distinctive structures made the classification task easier and allowed for a high classification accuracy at a  point density on the order of 1~$\mathrm{point}/\mathrm{m}^2$. 

Importantly, multiple studies have shown that radiometric features, such as the intensity of the laser echoes or surface reflectance, have a major impact on the species classification accuracy \citep{orka2009classifying, suratno2009tree, lin2016comprehensive, yu2017single, axelsson2018exploring,  shi2018tree, hakula2023individual}. For example, \citet{yu2017single} reported that the addition of single-channel intensity features improved the overall classification accuracy from 76.0\% to 85.9\%, while the corresponding improvement in  \citet{hakula2023individual} was from 73.2\% to 86.6\%. Further improvements to the classification accuracy may be achieved by incorporating spectral information, that is, using multiple wavelengths. This can be achieved either using multispectral ALS comprising multiple lidar channels at different wavelengths \citep[e.g.,][]{yu2017single, axelsson2018exploring, kukkonen2019multispectral,amiri2019classification, 
hakula2023individual, wang2024individual, taher2025} or combining passive multispectral or hyperspectral sensors with ALS \citep[e.g.,][]{dalponte2012tree,orka2013characterizing, deng2016comparison, shi2018tree, mayra2021tree}. 
Multispectral ALS has been shown to increase the overall classification accuracy compared to single-channel ALS from 66.5\% to 76.5\% in~\citet{axelsson2018exploring}, from 85.0\% to 88.2\% in~\citet{kukkonen2019multispectral}, from 86.6\% to 90.8\% in \citet{hakula2023individual}, and from 85.9\% to 87.9\%  in \citet{taher2025}. Similarly, a combination of passive imaging sensors and ALS has been demonstrated to improve the classification accuracy compared to single-channel ALS from 74.4\% to 87.0\% in \citet{orka2013characterizing} and from 65.1\% to 83.7\% in \citet{shi2018tree}.

In addition to data-related improvements, recent advancements in deep learning have led to the development of improved species classification methods outperforming traditional shallow machine learning methods both on terrestrial laser scanning (TLS) data \citep{xi2020see, liu2021tree} and on ALS data \citep{marinelli2022approach, taher2025}. Deep learning approaches for species classification can be divided into image-based 2D methods \citep[e.g.,][]{seidel2021predicting, marinelli2022approach, Detailview2024, korkeala2025} and point-based 3D methods \citep[e.g.,][]{xi2020see, liu2021tree, lin2024pctrees}. Image-based approaches project the point cloud into multiple 2D views that are given as input to convolutional neural networks, such as ResNet \citep{he2016deep} or YOLO \citep{Yolov82023}. On the other hand, point-based approaches take a point cloud with potential additional features as their input. Common point-based neural network architectures include PointNet++ \citep{qi2017pointnet++} and the point transformer \citep{zhao2021point}, for example. Importantly, deep learning approaches require a sufficiently large training dataset to outperform traditional machine learning models. Recently, \citet{taher2025} showed that the error of species classification follows a power law as a function of the training set size both for a point transformer model and a random forest classifier. The error of the deep learning model was shown to decrease faster with increasing training set size, especially for minority tree species. The crossover in the accuracy of deep learning and machine learning models was observed between 100 and 1000 training trees depending on the point cloud dataset.

Figure~\ref{fig: previous literature} summarizes the species classification performance achieved in previous ALS studies and highlights the best results achieved by \citet{taher2025} who used the MS-ALS-SPECIES dataset published with the present paper. We list the previous studies considered in  the analysis in Appendix \ref{ap: list of previous studies}.  
Figure~\ref{fig: previous literature}(a) shows the median overall accuracy of species classification  as a function of the number of tree species in previous studies using single-channel ALS, multispectral ALS, or ALS with passive multispectral/hyperspectral sensors. 
As expected, the overall accuracy generally decreases with increasing number of species. Furthermore, the median overall accuracy is slightly higher in studies using a combination of ALS and passive hyperspectral sensors compared to studies using single-channel ALS or multispectral ALS. However, a comparison between data acquisition types is not straightforward due to varying forest conditions and training set sizes across studies. 
Importantly, the multispectral ALS dataset published with the present article enabled \citet{taher2025} to achieve a state-of-the-art classification accuracy considering the  number of species included in the study.   \citet{taher2025} reached an overall accuracy of 92.0\% across 9 species  using 5000 training tree segments on the HeliALS dataset, which rivals the median accuracy reached in previous studies with 2--3 species. 

We also study trends in the overall accuracy reported in previous multispectral ALS studies as a function of point density and training set size in Figs.~\ref{fig: previous literature}(b) and (c), respectively. In general, the previous studies seem to be consistent with the expectation that the overall accuracy increases with increasing point density and training set size. However, it is difficult to observe clean trends, because conditions vary across studies. On the MS-ALS-SPECIES dataset published with the present article, the highest achieved overall accuracy was 92.0\% on the HeliALS dataset with a mean point density of 1300~$\mathrm{pts}/\mathrm{m}^2$. This is significantly higher than the accuracy of 83.7\% reached on the sparser Optech Titan dataset with a mean point density of 35~$\mathrm{pts}/\mathrm{m}^2$. Notably, recent studies using deep learning classifiers \citep{wang2024individual,taher2025} achieved significantly higher overall accuracies than previous shallow-machine-learning-based multispectral ALS studies with a comparable number of species. 

Since deep learning methods benefit from a large training set size, open laser scanning datasets are needed to foster further development of species classification methods and to enable systematic benchmarking of novel methods to previous ones. Notable previous open lidar datasets for species classification at the individual tree level include, e.g.,  \citet{graves_2020_dataset}, \citet{weiser2022dataset}, \citet{puliti2025benchmarking}, \citet{tockner2025tree}, and \citet{korkeala_2025_17639338}. \citet{graves_2020_dataset} provide single-channel airborne lidar data, high-resolution aerial orthorectified images, and hyperspectral surface reflectance measurements on three study areas located in three separate ecoclimatic domains in the United States. The dataset includes 20 species, and it is divided into a training set of 1057 individual trees and a test set of 585 trees. The ground-truth species labels of individual trees are represented as bounding boxes. The dataset was originally released for a data science competition focusing on tree species classification \citep{graves2023data}.

\citet{weiser2022dataset} provide single-channel ALS data, uncrewed aerial vehicle (UAV) laser scanning data (ULS) and TLS data for 1491 individual trees with ground-truth species labels on 12 forest plots in Germany. The dataset is introduced in a data paper \citep{weiser2022individual} and comprises 22 species representing a typical mixed forest in Central Europe. The mean point density  is 72.5~$\mathrm{points}/\mathrm{m}^2$ on the ALS data and on the order of 1000~$\mathrm{points}/\mathrm{m}^2$ on the ULS data.  \citet{tockner_2023_10035928} present a handheld laser scanning dataset of 6963 individual trees representing 9 species on 16 forest plots in Austria. The plots cover four climatic regions in Central Europe ranging from a dry oak--pine forest to a mountainous spruce--fir--larch forest. The point clouds were collected with the ZEB-Horizon scanner (GeoSLAM Ltd., Nottingham, UK) and they had a mean point density of $2.5 \times 10^4~\mathrm{pts}/\mathrm{m}^2$.  The dataset is further described in \citet{tockner2025tree}.  \citet{korkeala_2025_17639338} provide a handheld laser scanning dataset  comprising 1915 labeled trees of 8 species on a peri-urban study area in southern Finland in the boreal forest zone. The dataset was collected with the FARO Orbis scanner and it was used in a related publication to demonstrate a sensor-agnostic deep learning method NormalView for tree species classification \citep{korkeala2025}.

\begin{figure*}[ht]
    \centering
    \includegraphics[width=0.95\textwidth]{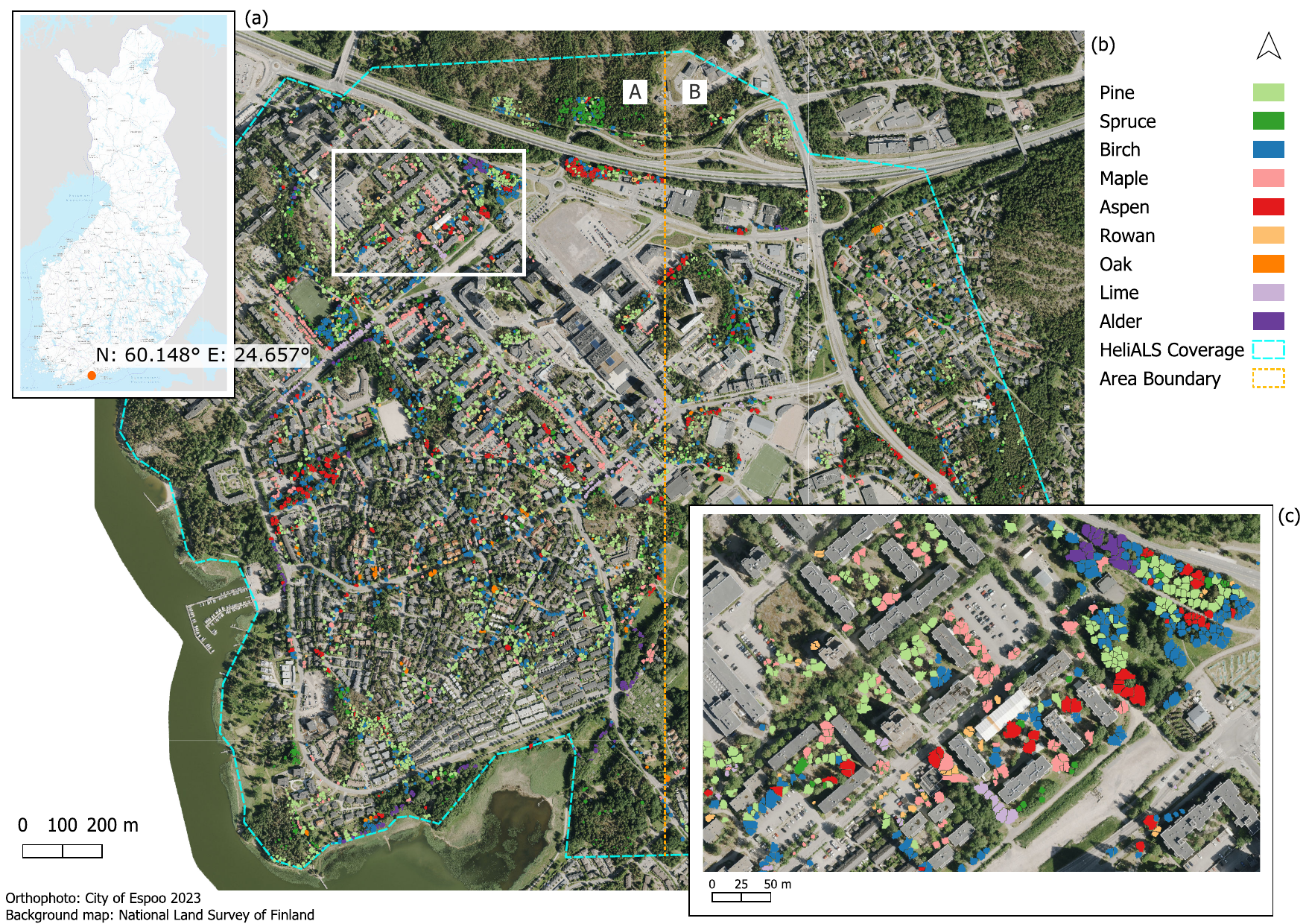}
    \caption{Study area of the  MS-ALS-SPECIES dataset. (a) The MS-ALS-SPECIES dataset is based on a study area located in Espoonlahti, southern Finland. (b) Orthophoto of the study area overlaid by tree segments with a reference species label. The light blue dashed line shows the boundary of the HeliALS point cloud data. The study area was divided into two areas A and B (orange dashed line), both of which were utilized for the open dataset published with this work. (c) A close-up of the orthophoto in (b) to an example suburban area within the study area. Figure adapted from \citet{taher2025}.}
    \label{fig:test_site}
\end{figure*}

Currently, the FOR-species20K dataset by \citet{puliti2025benchmarking} represents the largest open lidar dataset for species classification of individual trees. As the name suggests, the dataset includes over 20 000 trees of 33 species located in various climate and forest zones, including Mediterranean, temperate, and boreal biogeographic regions in Europe as well as scattered data on other continents. The dataset comprises  TLS, MLS, and ULS data, though the majority of the data has been collected by TLS. 

Importantly, all of the above referenced open datasets have been collected using single-channel lidar systems, and for example, the FOR-species20K does not include any intensity or reflectance information. Despite the potential of multispectral lidar technology, there are no previous open  multispectral lidar datasets for tree species classification. Thus, the MS-ALS-SPECIES dataset published with this article represents the first open multispectral lidar dataset for tree species classification. As described in more detail in Secs.~\ref{sec: ALS acquisition} and \ref{sec: open dataset}, MS-ALS-SPECIES dataset includes three-wavelength ALS data with point-level multispectral features for two airborne laser scanning systems at two distinct point densities: 35 $\mathrm{pts}/\mathrm{m}^2$ for the Optech Titan system and 1300 $\mathrm{pts}/\mathrm{m}^2$ for the HeliALS system. The dataset comprises 6326 tree segments representing 9 species, including 2 coniferous and 7 deciduous species in the hemiboreal forest zone. As the first multispectral ALS dataset with species information, the MS-ALS-SPECIES dataset is expected to foster research on deep learning methods for tree species classification using multispectral data.

\section{Study area, crowdsourcing application, and field data collection}
\label{sec:dataset}

In this section, we present the study area in Sec.~\ref{sec: test site}, the crowdsourcing application for efficient and scalable field reference data collection in Sec.~\ref{sec:crowdsourcing application}, and the details of the field data collection campaign in Sec.~\ref{sec:field_data_collection}.

\subsection{Study area}
\label{sec: test site}

The study area is located in Espoonlahti, Espoo, Finland, approximately 20 km west of Helsinki, with a center point at approximately 60.1462°N 24.6587°E. As shown in Fig.~\ref{fig:test_site}, the site represents a peri-urban area, which is partially covered by forests and partially by suburban built environment. The area is a mix of high-rise and low-rise residential zones, public buildings, a sports park, and recreational and non-recreational forests. The forests are typically sparser pine-dominated rocky forests or denser mixed forests containing both coniferous and deciduous trees. The site was chosen for its diverse tree species, which include a relatively high number of both naturally occurring and planted trees in contrast to typical managed boreal forests in Finland dominated by pine, spruce, and birch trees. Tree species on the site include: 

\begin{figure*}[ht]
    \centering
    \includegraphics[width=0.8\textwidth]{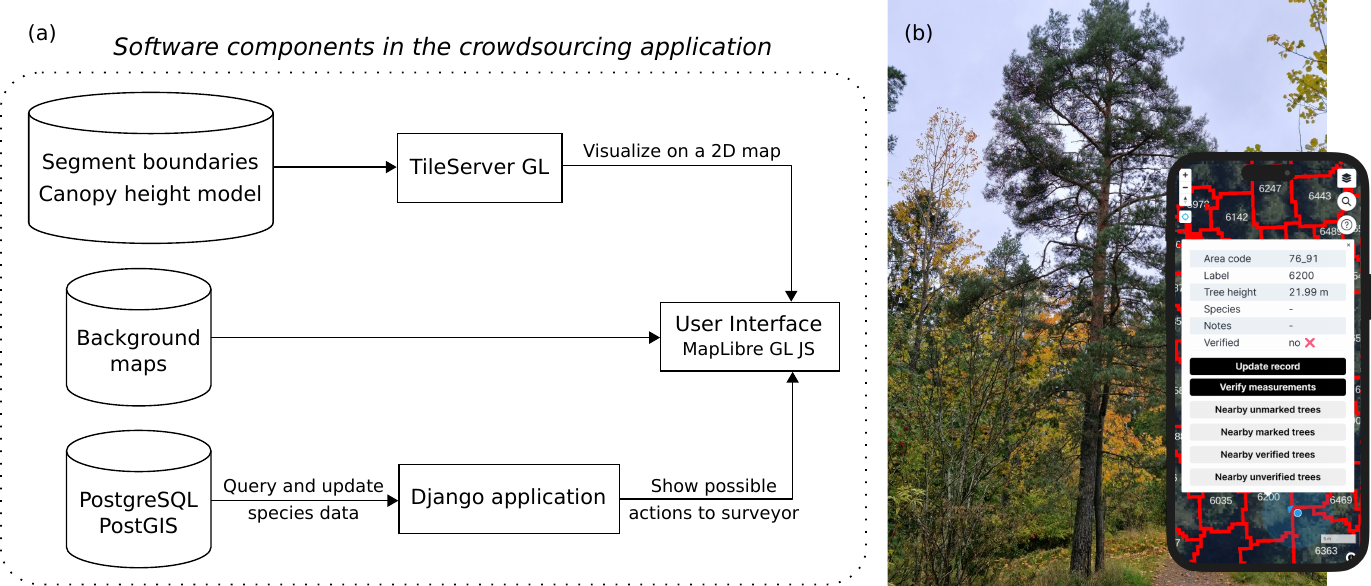}
    \caption{Crowdsourcing application for scalable collection of high-quality field reference data on tree species. (a) The software architecture of the crowdsourcing application developed to enable scalable ground truth collection for tree species classification. The backend of the crowdsourcing application stores boundaries of tree segments, a canopy height model and a database of tree attributes, including the species collected in the field. This information is visualized for the surveyor on top of third-party background maps in a user interface that allows specific actions, such as adding, modifying or verifying the species label of each tree segment. (b) Schematic illustration of field data collection using the crowdsourcing application. The photo shows an example tree in the field, while the user interface of the application presents the location of the surveyor on the background map with respect to the segment boundaries, known attributes of the tree, and possible actions for the surveyor. The background map is the true orthophoto map from the \citet{espoo_open_data}.}
    \label{fig:crowdsourcing_tool_architecture}
\end{figure*}

\begin{itemize}
    \item Mainly naturally occurring tree species: pine (\textit{Pinus sylvestris}), spruce (\textit{Picea sp.}), birch (\textit{Betula sp.}), aspen (\textit{Populus tremula}), rowan (\textit{Sorbus sp.}), and alder (\textit{Alnus sp.}). These are found in both forests and the built environment; in the latter case, they may also be planted trees.
    \item Mainly planted species: maple (\textit{Acer platanoides}), oak (\textit{Quercus robur}), and linden (\textit{Tilia sp.}), primarily found in the built environment.
\end{itemize}

\subsection{Crowdsourcing application for ground-truth collection}
\label{sec:crowdsourcing application}

Methods for scalable and efficient field reference collection are increasingly needed to enable operational-scale species classification of individual trees since the accuracy of deep learning classification methods improves with increasing training set size following a power law.
To this end, a crowdsourcing application was developed to facilitate scalable annotation of tree species labels of tree segments in the field. The crowdsourcing tool, accessible via standard web browsers, utilizes GNSS positioning and features a 2D geospatial visualization of the trees around the surveyor. 
Within the application interface, the trees are represented by their pre-delineated segment boundaries derived from the laser scanning data as explained in Sec.~\ref{sec: ALS acquisition}. For visualization purposes, the segment boundaries are overlaid onto a background map. Owing to GNSS positioning, the developed application helps surveyors to locate themselves in the field with respect to the segment boundaries and enables several surveyors to annotate species labels simultaneously. This allows scalable and efficient collection of ground truth data in areas where the GNSS signal is not fully obstructed. Thus, the crowdsourcing application provides a promising approach for collecting ground truth data for large-scale tree species datasets,
benefiting especially deep-learning-based species classification methods.

The architecture of the crowdsourcing application is illustrated in Fig.~\ref{fig:crowdsourcing_tool_architecture}(a). The application consists of the user interface, backend services, and a PostgreSQL database \citep{PostgreSQL}. The user interface is built with the React library \citep{React} making it easy to split the code of the user interface into modular, reusable components. The 2D geospatial visualization of the study area is rendered using MapLibre GL JS library \citep{MapLibreGLJS} and react-map-gl \citep{ReactMapGL} which provides React components for MapLibre GL JS. The user interface relies on external background maps and the user can choose from a true orthophoto map from the \citet{espoo_open_data} and a topographic map from \citet{nls_vector_tiles}. There are two main backend services in the platform: TileServer GL \citep{TileServerGL} and a self-developed Django application \citep{Django}. TileServer GL is utilized for providing the pre-delineated segment boundaries and the canopy height model (CHM) of the study area for the map view of the user interface. The self-developed Django application interacts with the PostgreSQL database and provides an application programming interface (API) for the user interface to retrieve and update information about the tree segments. The PostgreSQL database stores the coordinates and heights of the tree segments in addition to the tree species and textual annotations. PostGIS database extension \citep{PostGIS} is utilized to add support for storing, indexing, and querying geospatial data. The database is installed and run on a separate virtual machine, whereas the backend services are run using Docker \citep{Docker}.

\subsection{Field data collection}
\label{sec:field_data_collection}

\begin{figure*}[ht]
    \centering
    \includegraphics[width=\textwidth]{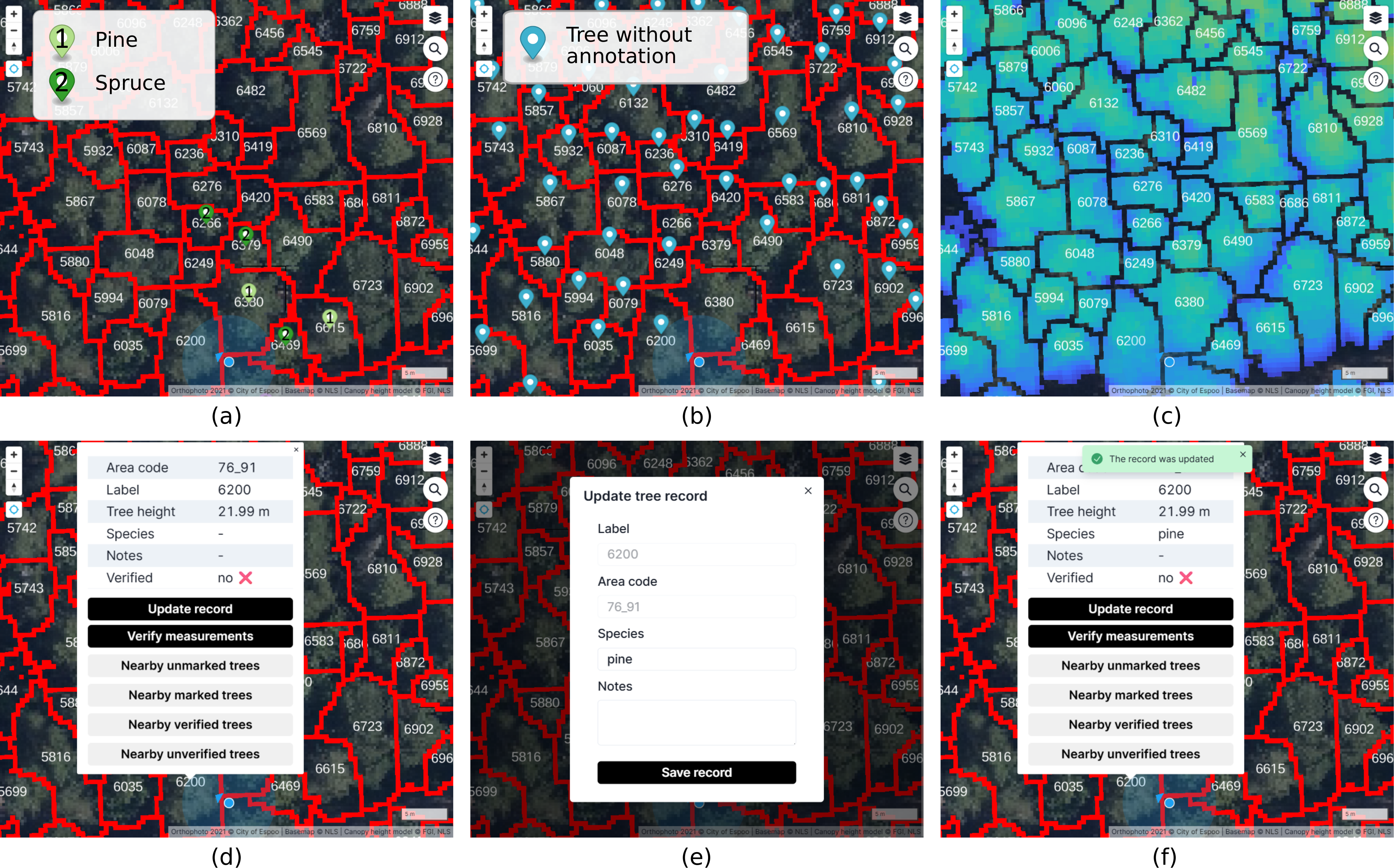}
    \caption{The workflow for annotating species labels in the field using the crowdsourcing application. The platform can highlight (a) tree segments which have already been annotated and (b) tree segments in need of annotation. The location of the surveyor is shown by the blue circle. (c) In addition to GNSS positioning and a true orthophoto background map, the surveyors can use the canopy height model of the study area to help positioning in the field and making sure the segment on the map visualization matches the one in the field. (d) When clicking a segment, the platform shows the recorded information about the segment and possible actions for the surveyor.  (e) The surveyor annotates the species of a segment and can record supplementary textual annotations if needed. (f) The information about the species label has been updated after the species was annotated. The label has not yet been verified. The background map in all panels is the true orthophoto map from the \citet{espoo_open_data}.}
    \label{fig:crowdsourcing_tool}
\end{figure*}

During the autumn of 2023 and the summer of 2024, field surveyors used the crowdsourcing application to annotate the tree segments with ground-truth species labels and to write supplementary textual annotations if needed. Figures~\ref{fig:crowdsourcing_tool_architecture}(b) and~\ref{fig:crowdsourcing_tool}(a)--(f) depict our workflow for annotating species labels in the field using the crowdsourcing application. Before visiting the study area, a field surveyor identified an area where most of the tree segments had not yet been annotated. The application allowed visualizing both trees with an existing ground-truth annotation as shown in Fig.~\ref{fig:crowdsourcing_tool}(a) as well as trees in need of annotation as shown in Fig.~\ref{fig:crowdsourcing_tool}(b). 
The field surveyors utilized GNSS positioning through the application to locate themselves in the field. To aid positioning under dense canopies  with limited GNSS signal reception, it was also possible to select a background map based on a true orthophoto map from the \cite{espoo_open_data} as shown in Figs.~\ref{fig:crowdsourcing_tool}(a)--(b).
To verify that the tree segment on the crowdsourcing platform corresponded to the nearest tree in the field, the surveyor could also utilize information about nearby trees, including their heights and previously annotated species labels. Therefore, a background map showing the CHM  was also available, see Fig.~\ref{fig:crowdsourcing_tool}(c). Because the CHM and the tree segments were derived from the Optech Titan data collected on June 14, 2016 (see Sec.~\ref{sec: optech data}), the surveyor needed to be careful to take into account potential changes in the environment between the ALS data acquisition and the date of field measurements. 
Figure~\ref{fig:crowdsourcing_tool}(d) shows the information of the nearest tree, such as an estimated height of 21.99 m in 2016, together with possible actions for the surveyor. 
After confirming that the segment on the crowdsourcing platform matched the tree next to the surveyor, the surveyor recorded the species in the database of the platform, see Fig.~\ref{fig:crowdsourcing_tool}(e). 
After recording the species of the current segment, the information of the segment was updated as shown in Fig.~\ref{fig:crowdsourcing_tool}(f), and the surveyor would continue the same process to annotate other  tree segments. 

The crowdsourcing platform was also used as an on-site verification tool that enabled independent confirmation of species labels after the first round of data annotations. To help determine which segments to prioritize in the verification of the dataset, we computed for each tree segment the proportion of  classification methods participating in \citet{taher2025} that predicted a  species label different from the field-annotated label. Based on the results, we prepared a map illustrating the proportion of conflicting predictions for each tree to guide the verification in the field. During the second round of field visits, we managed to identify and fix some errors in the ground-truth species labels that had been caused by positioning errors and rarer tree species incorrectly labeled as another species. In total, 28.5\% of the labeled segments were revisited and  verified using the crowdsourcing platform during the second round of field visits.

Even though the crowdsourcing application made the field reference data collection significantly easier, the task still required practice in recognizing the tree species and in positioning oneself on the map in complicated environments with some uncertainty in GNSS positioning. 
Therefore, all segments were annotated by employees of the FGI to ensure the quality of the field reference data. Most trees were labeled in the field, but some clear instances of spruces were annotated using the true orthophoto map with 5-cm resolution due to the unique shape of the spruces in top view images. In addition, some planted deciduous roadside trees were annotated using the open tree database of the \cite{espoo_open_data}.

\section{Multispectral ALS data collection}
\label{sec: ALS acquisition}

Multispectral airborne laser scanning data was collected on the study area using two distinct systems---the Optech Titan system providing a mean point density of 35 $\mathrm{pts}/\mathrm{m}^2$  and the FGI-developed HeliALS system resulting in a mean point density of 1300 $\mathrm{pts}/\mathrm{m}^2$. The point density of the Optech Titan system is representative of typical city-wide scanning campaigns, whereas the high-density HeliALS dataset (HeliALS) is designed for automating plot- and real-estate-level field inventories and expected to be representative of future UAV-borne multispectral laser scanning data. We have previously described the collection of the multispectral ALS data in our benchmarking study on tree species classification  \citep{taher2025}. For completeness, we provide here the key details of the data collection.

\subsection{Optech Titan}
\label{sec: optech data}

 The Optech Titan instrument is a three-channel multispectral ALS system from Teledyne Optech, debuting in 2014.  The Optech Titan measurements were carried out in cooperation with TerraTec Oy (Helsinki, Finland) on June 14, 2016, when the trees were in full leaf \citep{karila2019effect}.  The system was operated from a fixed-wing aircraft flying at an altitude of 700~m, which resulted in three separate point clouds with the following mean point densities:
 \begin{itemize}
   \item infrared ($\lambda=1550$ nm) with 11~pts/$\mathrm{m}^2$,
   \item near-infrared ($\lambda=1064$ nm) with 13~pts/$\mathrm{m}^2$, and
   \item green ($\lambda=532$ nm) with 11~pts/$\mathrm{m}^2$. 
\end{itemize}
Here, the reported densities were estimated after preprocessing steps, such as cutting points of overlapping flight lines. Table~\ref{tab: Optech Titan specs} lists more detailed specifications of the instrument and Espoonlahti measurements.

\begin{table}[ht!]
    \centering
    \caption{Specifications of the Optech Titan system mounted on a fixed-wing aircraft.} 
    \label{tab: Optech Titan specs}
    \resizebox{0.5\textwidth}{!}{\begin{tabular}{llll}
        \toprule
        Flight parameters & Value \\
        \hline
         Altitude AGL (m) &  700\\
         Flight speed (m/s) & $\sim$70 \\
         Strip width  (m) &  510 \\
         Lateral overlap & 30\%  \\
         \toprule
        Channel-specific Characteristics & Channel 1 & Channel 2 & Channel 3 \\
         \hline
         Laser wavelength (nm) & 1550 & 1064& 532 \\
         Beam divergence  (mrad) & 0.35 & 0.35 & 0.7 \\
         Beam diameter at ground (cm) & 25& 25& 50\\
         Direction (degrees) & 3.5 forward & nadir & 7 forward \\
         Field of view (degrees) & $40$ & $40$& $40$  \\
         Pulse repetition rate (kHz) &  300 & 300  & 300 \\
         Resulting point density$^*$  ($\mathrm{pts}/\mathrm{m}^2$) & 11&  13 &  11\\
         \bottomrule
         \multicolumn{4}{l}{\footnotesize{$^*$ After cutting points of overlapping flight lines.}}
    \end{tabular}}
\end{table}

In addition to basic preprocessing of the laser scanning data by TerraTec Oy, we performed some additional steps, including a range ($R$) correction for the intensity according to $R^2$ \citep{matikainen2017object}, as well as removal of overlapping points from different flight lines and cross lines. Thus, the resulting  point clouds include multispectral information in the form of range-corrected intensities for each of the three channels. However, reflectance information is not available.

The point cloud was further semantically segmented into ground, vegetation, building, and noise points using LAStools software (rapidlasso GmbH, Germany). All points with 3 or fewer other points in their surrounding within a voxel of 6 by 6 by 6 meters were regarded as noise points. %
Furthermore, points below ground and higher than 50 meters above ground were also classified as noise. In ground filtering, a step size of 25 meters was used  to remove most buildings. For other parameters, default values in LAStools were used for the preprocessing steps. A digital terrain model was further created using the ground points, which enabled point cloud normalization by subtracting the ground elevation from the \(z\)-coordinates of each laser point.

Trees were automatically segmented from the Optech Titan data using a varying-window watershed method, first proposed in \citep{kaartinen2012international}. The selection of the method was motivated by the relatively large number of deciduous trees on the study area. First, a canopy height model (CHM) with a pixel size of 0.5 m was created using the first returns of the vegetation points from channels 1 and 2. The CHM was smoothed using a Gaussian filter with a window size varying according to the height of the target as follows: 3 pixels for $h\in [0, 7)$ m, 5 pixels for $h \in [7, 20)$ m, 7 pixels for $h \in [20, 30)$ m, and 9 pixels for $h > 30$ m. Subsequently, tree tops were detected as local maxima of the smoothed CHM, and the boundaries of tree segments were determined using the marker-controlled watershed algorithm with the tree tops as seeds \citep{kaartinen2012international}. The segment boundaries were also directly transferred to the HeliALS dataset to carry out tree segmentation. This ensured the correspondence of the tree segments across the two laser scanning datasets.

\subsection{HeliALS}
\label{sec: helials data}

The HeliALS system, first demonstrated in \citet{hakula2023individual}, consisted of the following sensors mounted on a helicopter:
\begin{itemize}
    \item three Riegl drone laser scanners operating at distinct wavelengths and resulting in different mean point densities:
    \begin{itemize}
             \item VUX-1HA ($\lambda=1550$ nm, 581~pts/$\mathrm{m}^2$), 
             \item miniVUX-1DL ($\lambda=905$ nm, 175~pts/$\mathrm{m}^2$),
             \item VQ-840-G ($\lambda=532$ nm, 519~pts/$\mathrm{m}^2$). 
    \end{itemize}
    \item positioning hardware consisting of a NovAtel (LITEF) ISA-100C inertial measurement unit (IMU), a NovAtel PwrPak7 Global Navigation Satellite System (GNSS) receiver, and a NovAtel (Vexxis) GNSS-850 antenna.
\end{itemize}
The HeliALS dataset was collected on July 20 and 28, 2023 under leaf-on conditions at an altitude of 100 m and a speed of 14 m/s.  More detailed specifications of the laser scanning system and Espoonlahti measurements are listed in Table~\ref{tab: HeliALS specs}. 

\begin{table}[ht!]
    \centering
    \caption{Specifications of the helicopter-borne HeliALS system.} 
    \label{tab: HeliALS specs}
   \resizebox{0.5\textwidth}{!}{
   \begin{tabular}{llll}
        \toprule
        Flight parameters & Value \\
        \hline
         Altitude AGL (m)  &  100 \\
         Flight speed (m/s) & 14\\
         \toprule
        Channel characteristics & 1 & 2 & 3 \\
         \hline
         Scanner name & VUX-1HA & miniVUX-1DL & VQ-840-G \\
         Laser wavelength (nm) & 1550 & 905 & 532 \\
         Beam divergence (mrad) & 0.5 & \(0.5 \times 1.6\) & 1 \\
         
         Footprint at ground (cm) & 5 & \( 5 \times 16 \) &  10 \\
         Range accuracy (mm) & 5 &  15 &  20 \\
         Direction (degrees) & 15 & Nadir &  Nadir\\
         Field of view (degrees) & 180 & 46 & 40 \\
         Scanning geometry & 360\textdegree \ ring & Conical & Conical \\
         Pulse repetition rate (kHz) & 1017 & 100 & 200 \\
         Point density (pts/$\mathrm{m}^2$) & 581 & 175 & 519 \\
         \bottomrule
    \end{tabular}}
\end{table}

The HeliALS data at three wavelengths were processed into point clouds using RiProcess software (version 1.9.0, Riegl GmbH, Austria). Waypoint Inertial Explorer (8.90, NovAtel Inc., Canada) was employed to estimate GNSS-IMU trajectories with the help of a virtual base station from the Trimnet service (Geotrim Oy, Finland) and tightly coupled kinematic post-processing. Boresight calibrations between the IMU and each scanner were computed with RiProcess functionalities that use planar surfaces to estimate the rotation angles. 
The resulting point cloud combining the three channels had a mean point density of approximately 1300~$\mathrm{pts}/\mathrm{m}^2$.

The HeliALS point cloud was further processed by first identifying noise points using a statistical outlier filter \citep{rusu2007towards}, followed by ground classification with a cloth simulation filter \citep{zhang2016easy}. All processing steps were implemented using the Point Data Abstraction Library (PDAL, version 2.6.2, \citet{PDAL}) with python bindings (version 3.2.3). Noise points were excluded from the subsequent ground-classification stage. The parameters of the statistical outlier filter were tuned through visual inspection, and were set to: number of nearest neighbors = 6 and sigma multiplier = 2.0. Similarly, the cloth simulation filter parameters were optimized through visual verification and set as follows: resolution = 1.0, threshold = 1.5, smoothing = enabled, step = 0.65, rigidness = 1.0 and number of iterations = 500.
As the final processing step, the point cloud was segmented using the boundaries of the tree segments obtained on the Optech Titan data.

\section{Open multispectral ALS dataset}
\label{sec: open dataset}

In this section, we present the open dataset published in Zenodo~\citep{hyyppa_2025_zenodo} with the goal to provide useful background information for future studies using the dataset.  Section \ref{sec: ground-truth reference data} describes the structure of the ground-truth reference data, while the details of the point cloud data are presented in Section~\ref{sec: point cloud data}.

\subsection{Ground-truth reference data}
\label{sec: ground-truth reference data}

\begin{figure*}[ht]
    \centering
    \includegraphics[width=0.74\textwidth]{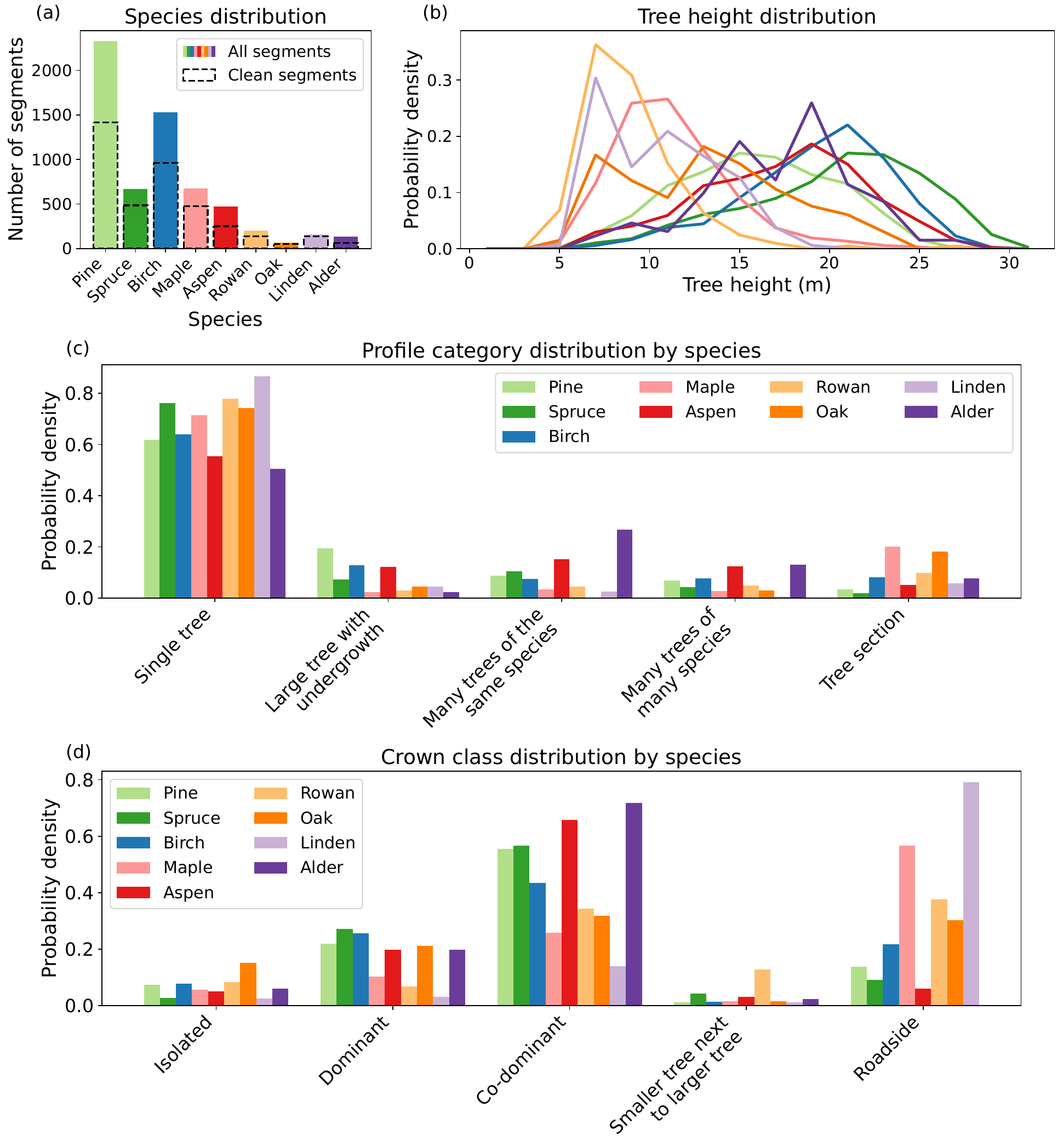}
    \cprotect\caption{ Statistics of the open MS-ALS-SPECIES dataset based on the final state of the reference data (\verb|final-segments-with-species.csv|). (a) Tree species distribution showing the number of segments in each species class for the entire dataset (colored bars) and for `clean segments' (dashed bars). (b) Tree height distribution for each of the nine tree species, with the same color scheme as in (a). (c) Profile category distribution by species (see Sec.~\ref{sec: ground-truth reference data}). (d) Crown class distribution by species (see Sec.~\ref{sec: ground-truth reference data}).}
    \label{fig:statistics}
\end{figure*}

The open MS-ALS-SPECIES dataset contains 6326 segment-level tree point clouds for both HeliALS and Optech Titan datasets with the corresponding ground-truth species labels.  The ground-truth species labels are available as CSV files containing the following attributes:  \texttt{test\_site\_name}, \texttt{segment\_id}, \texttt{height}, \texttt{species\_code}, \texttt{profile\_category}, \texttt{profile\_category\_name}, \texttt{crown\_class}, and \texttt{crown\_class\_name}. The attribute \texttt{test\_site\_name} takes the value A or B, corresponding to the original division of the Optech Titan data into two separate areas, see Fig.~\ref{fig:test_site}.
The attribute \texttt{segment\_id} is the identifier of the segment and is unique within the area A or B containing the segment. The attribute \texttt{height} represents the height of the segment in meters from the ground level as estimated from the Optech Titan data collected in 2016. The attribute \texttt{species\_code} is the ground-truth species class assigned to the segment, see Sec.~\ref{sec:field_data_collection}. The values range from 1 to 9 corresponding to species in the following order: pine, spruce, birch, maple, aspen, rowan, oak, linden, and alder. The attributes \texttt{profile\_category} and \texttt{profile\_category\_name} are the identifier and descriptive name of profile categories assigned for each tree segment by manually inspecting the segmentation quality based on 2D projections of the HeliALS point cloud segments. The considered profile categories include  ‘Single tree’, ‘Large tree with undergrowth’, ‘Many trees of the same species’, ‘Many trees of many species’, and ‘Tree section’, see \citet{taher2025} and below for more details. The attributes 
\texttt{crown\_class} and \texttt{crown\_class\_name} are the labels and the descriptive names of the crown classes based on the local neighborhood of each tree segment. The considered crown classes include ‘Isolated’, ‘Dominant’, ‘Co-dominant’, ‘Smaller tree next to larger tree’, and ‘Roadside’ as detailed in \citet{taher2025} and below. 

 Two versions of the ground-truth species labels are publicly available as separate CSV files: \verb|final-segments-with-species.csv| containing the final state of the field reference data for 6238 tree segments and \verb|training-and-test-segments-with-species.csv| reflecting the original setting of the benchmarking competition in \cite{taher2025} with 6326 tree segments. In \verb|final-segments-with-species.csv|, we have excluded 88 segments from the original training set of the benchmarking competition (\verb|training-and-test-segments-with-species.csv|) due to non-ideal profile images, sparse point clouds, or building intersection within the segment boundaries. In addition, the species class of 18 segments from the training set of \verb|training-and-test-segments-with-species.csv| was corrected for \verb|final-segments-with-species.csv| due to errors detected and fixed during the on-site verification of the ground-truth species labels. To obtain new results comparable to the the results of the benchmarking competition \citep{taher2025}, readers are advised to use the species labels in \verb|training-and-test-segments-with-species.csv| containing 6326 tree segments. For this data, the CSV file contains an additional attribute \texttt{test\_set\_flag}, where values 0 and 1 correspond to segments in the training and test sets of the benchmarking contest, respectively. The original training set contains 1065 tree segments, while the test set contains 5261 segments. Otherwise, readers are advised to use \verb|final-segments-with-species.csv| containing 6238 tree segments reflecting the final state of the field reference data. 

Figure~\ref{fig:statistics} shows statistics of the species, height, segmentation quality, and crown class for the tree segments included in our dataset. First, Fig.~\ref{fig:statistics}(a) illustrates the distribution of tree species in our dataset.  In order from the most occurrences to the least, the species are pine, birch, maple, spruce, aspen, rowan, linden, alder, and oak. The species distribution is relatively imbalanced because some minority species are considerably rarer on the study area compared to the most common species.
We further show the distribution of tree height  for each of the nine species in Fig.~\ref{fig:statistics}(b). Importantly, the typical height varies strongly by species. For example, spruces and birches are generally large trees, whereas rowans and lindens are significantly smaller. 

Figure~\ref{fig:statistics}(c) shows the profile category distribution of the dataset by species. The segments were categorized based on profile images generated from the HeliALS data from a given direction. 
The vast majority of the tree segments belong to the `Single tree' profile category. Segments were assigned to this category if there was only one tree visible in the profile image and there was no visible cut-off indicating that the tree had been divided into multiple segments. Segments were assigned to the `Large tree with undergrowth' category, instead, if there were smaller trees visible in the profile image but their height was at most one third of the height of the tallest tree in the segment. If the segment contained at least two trees of the same species but no other species, the segment was categorized into the `Many trees of the same species' category. In the dataset, 30\% of alders belong to the `Many trees of the same species' category because alders typically grew in dense clusters containing a large number of alder trees, see Fig.~\ref{fig:aspen_alder}. Furthermore, `Many trees of many species' corresponds to segments with at least two trees of different species. For the segments belonging to the `Many trees of many species' category, the species label was determined based on the tallest tree in the segment. Segments, in which a single prominent tree species could not be identified, were removed from the dataset.  `Tree section' denotes a tree segment containing only a part of a single tree, for example, resulting from the over-segmentation of a wide tree canopy. 

The training set reflecting the original setting of the benchmarking competition in \cite{taher2025} (\verb|training-and-test-segments-with-species.csv|) also contains some segments with the profile categories `Building visible', `Sparse segment', `Unclear segment', and `No trees'. `Building visible' corresponds to segments whose boundaries intersect with a building. `Sparse segment' corresponds to segments whose point cloud in HeliALS data looked significantly sparser compared to other segments in the manual profile inspection. If the profile image of a segment looked unclear and could not be assigned to other categories, it was classified as `Unclear segment'. `No trees' corresponds to segments where no trees were visible in the profile inspection, possibly resulting from a tree being cut or a positioning error in the field, for example. None of these profile categories are present in the version of the dataset containing 6238 tree segments reflecting the final state of the field reference data.

Furthermore, Fig.~\ref{fig:statistics}(d) shows the crown class distribution of the dataset by species. Segments were considered `Isolated' if there were no trees within $8~\mathrm{m}$ radius or the trees within $8~\mathrm{m}$ radius were at most half of the height of the given segment. If a segment was the tallest within $8~\mathrm{m}$ radius but not `Isolated', it was considered `Dominant'. A segment was classified into the `Smaller tree next to larger tree' category if there was at least one other tree at least $5~\mathrm{m}$ taller within $6~\mathrm{m}$ radius from the segment. A tree segment was classified as `Roadside' if the tree was located in a built environment area such as a parking lot or street, in a clear line. Clusters of trees were not classified as roadside trees. The open tree database of the \cite{espoo_open_data} contains locations of roadside trees, which were also utilized in QGIS \citep{QGIS_software} to help classifying them. All other segments were classified into the `Co-dominant' crown class. Figure~\ref{fig:statistics}(d) shows that most of the segments belong to the crown classes "Co-dominant" or "Roadside".

In section ~\ref{sec:results and discussion}, we further investigate a group of high-quality tree segments named as `clean segments' to assess whether the results of the original benchmarking study in \citet{taher2025} are changed by excluding non-ideal tree segments. A tree segment is classified as a `clean segment' if it belongs to the profile category `Single tree' and does not belong to the crown category `Smaller tree next to larger tree'. This was done to separate high-quality segments representing taller, isolated, or dominant trees from  low-quality segments that might contain parts of multiple trees due to being located in a dense cluster of trees. See Fig.~\ref{fig:statistics}(a) for the species distribution of `clean segments'.

\subsection{Point cloud data}
\label{sec: point cloud data}

\begin{figure*}[ht]
    \centering
    \includegraphics[width=0.9\textwidth]{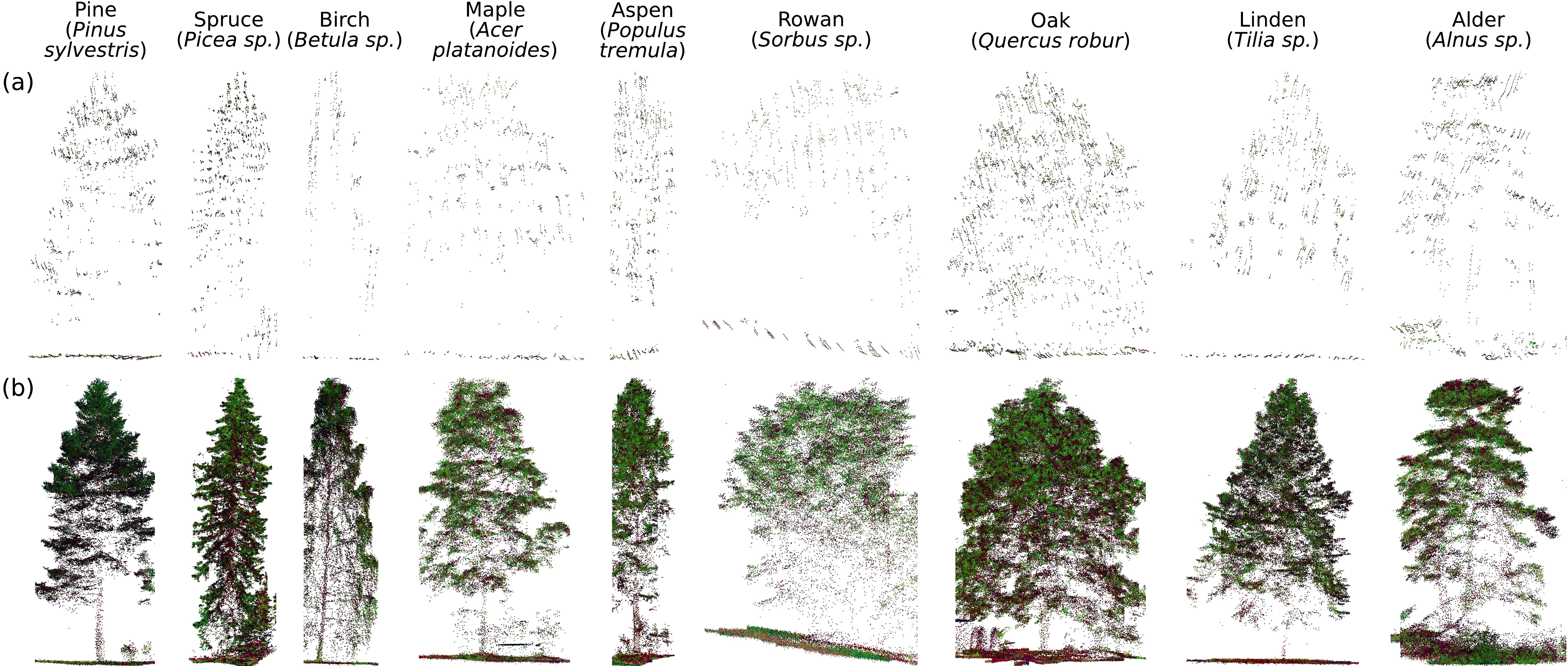}
    \caption{Orthographic projections of multispectral point clouds of tree segments in (a) Optech Titan and (b) HeliALS datasets. The top and bottom rows depict the same tree instance for each species sample. Points are colored by nearest-neighbor interpolated, transformed, and gamma-corrected intensity values. The scale varies across species. }
    \label{fig:helials_optech_point_clouds}
\end{figure*}

In the open MS-ALS-SPECIES dataset, point clouds are delivered in LAS version 1.2 using point format 3.
Individual tree segments are stored as separate LAS files, and the datasets originating from the HeliALS and Optech Titan systems are organized into distinct file collections.
Figure ~\ref{fig:helials_optech_point_clouds} illustrates the point density differences and the extent of laser-pulse penetration through the canopy between the Optech Titan and HeliALS systems.

\begin{figure*}[ht]
    \centering
    \includegraphics[width=0.9\textwidth]{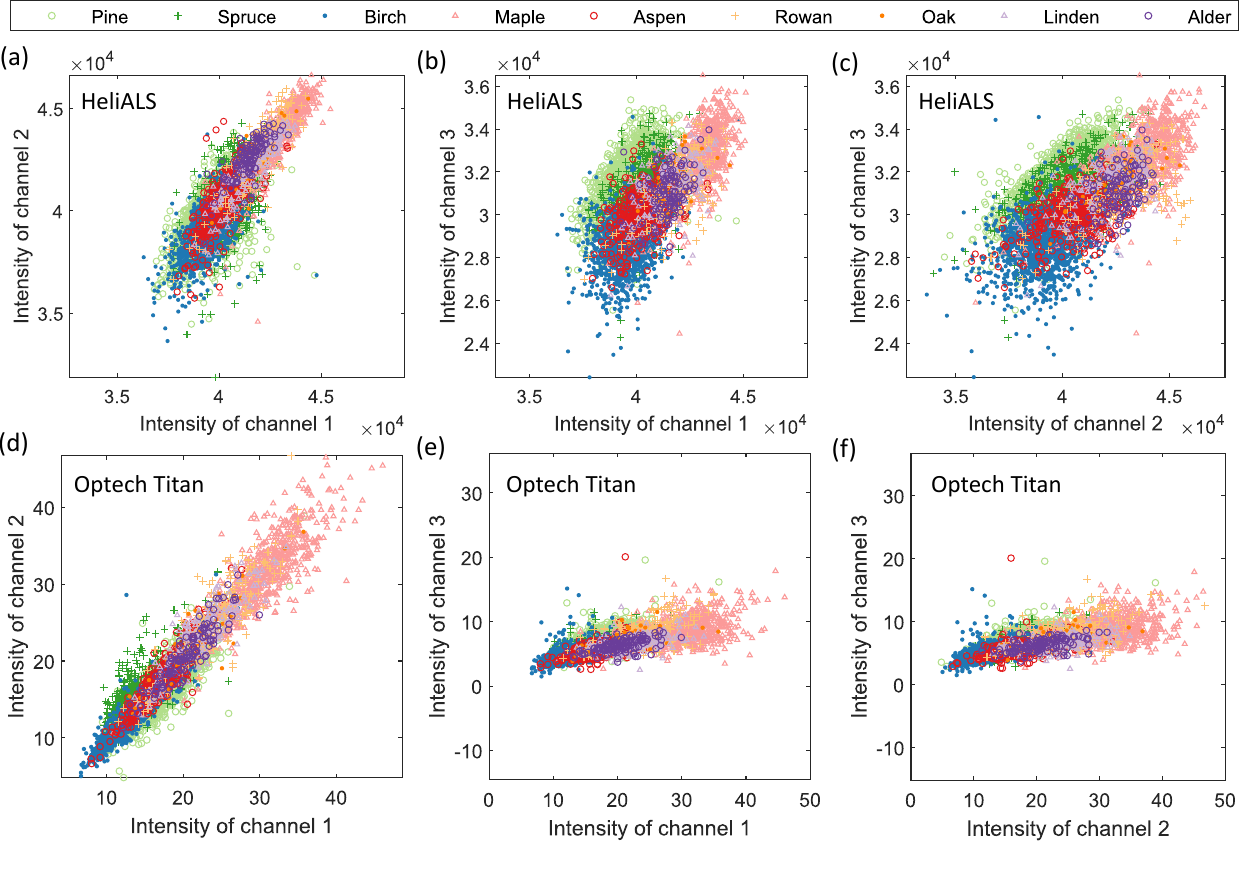}
    \caption{Scatter plots for the mean intensity of the three channels for each tree segment in the multispectral HeliALS and Optech Titan datasets. (a) Scatter plot of mean intensities of channels 1 and 2 in arbitrary units for each tree segment in the HeliALS dataset colored by ground-truth species. (b) Same as (a) but for channels 1 and 3. (c) Same as (b) but for channels 2 and 3. (d) Scatter plot of mean intensities of channels 1 and 2 in arbitrary units for each tree segment in the Optech Titan dataset colored by ground-truth species. (e) Same as (d) but for channels 1 and 3. (f) Same as (d) but for channels 2 and 3.}
    \label{fig:intensity scatter plot}
\end{figure*}

The HeliALS point clouds include the following attributes: $x$-, $y$-, and $z$-coordinates in the Finnish ETRS-TM35FIN coordinate system (with $z$-coordinate representing ellipsoidal height); 16-bit intensity; return number and number of returns per emitted pulse (up to seven returns); scan direction flag; edge of flight line indicator; classification into ground and noise points; scan angle rank; user data identifying the scanner from which the point originated (1 = VUX-1HA, 1550 nm; 2 = miniVUX-1DL, 905 nm; 3 = VQ-840-G, 532 nm); point source id referencing to sequential one-minute data acquisition blocks; gps time; RGB values derived from nearest-neighbor interpolation of multispectral intensity (search radius of 0.25 m); and additional radiometric attributes including amplitude, reflectance and echo deviation \citep[for further details, please refer to][]{pfennigbauer2010improving}.

For visualization, normalized 16-bit point-wise color values $I_{c,\text{norm}} \in \mathbb{R}$ were computed for each channel $c \in \{\text{R,G,B\}}$ as
\begin{equation*}
    I_{c,\text{norm}} = \left( \frac{\operatorname{clip}(I_{c} | L_{c, \min}, L_{c, \max}) - L_{c, \min}}{L_{c, \max} - L_{c, \min}} \right)^{\gamma} \times 2^{16} ,
\end{equation*}
where the gamma correction factor was set to $\gamma = 2.5$, and the channel specific clipping limits to: $\mathbf{L}_{\text{red}} = [30000, 50000]$, $\mathbf{L}_{\text{green}} = [32000, 45000]$ and $\mathbf{L}_{\text{blue}} = [15000, 55000]$. The clipping operation is defined as
\begin{equation*}
\operatorname{clip}(I_{c} | L_{c, \min}, L_{c, \max}) = \min (L_{c, \max} , \max (I_{c}, L_{c, \min})) \ .
\end{equation*}
Intensity values recorded by the VUX-1HA, miniVUX-1DL, and VQ-840-G sensors were mapped to the red, green and blue color channels, respectively. The parameters, and the mapping from the sensors to the color channels were chosen purely on the basis of visualization purposes. Therefore, it is not advisable to use the RGB color values in automated analysis instead of the original intensity channel values.

The Optech Titan point clouds contain the following attributes for each point: $x$-, $y$-, and $z$-coordinates in the ETRS-TM35FIN coordinate system (with $z$-coordinate exhibiting an approximate 10-20 cm offset relative to the N2000 height system); 12-bit intensity; return number and number of returns per pulse (up to four returns); scan direction flag; classification into ground, high vegetation, building and noise points; scan angle rank, user data indicating the scanner from which the point originated (1 = channel 1, 1550 nm; 2 = channel 2, 1064 nm; 3 = channel 3, 532 nm); point source id corresponding to the acquisition flight line; gps time; RGB values derived from nearest-neighbor interpolation of intensity (search radius of 1.0 m); and height above ground level (AGL).

Color mapping for visualization followed a three-stage procedure.
First, intensities $I_{c} \in \mathbb{R}$ for each channel $c \in \{\text{R,G,B\}}$ were clipped based on the channel-wise median and standard deviation
\begin{equation*}
    \tilde{I}_{c} = \max\bigl(0, \min\bigl(I_{c}, \operatorname{median}(\textbf{I}_{c}) + a \cdot \sigma(\textbf{I}_{c})\bigr)\bigr) \ ,
\end{equation*}
where the standard deviation multiplication factor was set to $a = 2$ which reduces the influence of extreme radiometric outliers, and $\textbf{I}_{c}$ denotes intensity values for channel $c$ for the set of all points in the unsegmented point cloud.
The full point cloud was used to ensure consistent color values among the tree segments. The global maximum clipped intensity value across all channels was then computed as
\begin{equation*}
    \tilde{I}_{\max} = \max_{c \in \{\text{R,G,B}\}} \left( \max(\tilde{\textbf{I}}_c) \right).
\end{equation*}
Finally, the clipped values were normalized and scaled to 16-bit depth as
\begin{equation*}
    I_{c,\text{norm}} = \frac{\tilde{I}_c}{\tilde{I}_{\text{max}}} \times (2^{16} - 1) \ ,
\end{equation*}
in order to obtain the the final processed point-wise color value $I_{c,\text{norm}}$ for channel $c$.
Intensity values from Channel 1 (1550 nm), channel 2 (1064 nm) and channel 3 (532 nm) were assigned to the red, green and blue color channels, respectively. Similarly to the HeliALS data, the Optech Titan color channel values should be used for visualization purposes only, and further automated analysis should be carried out using the original intensity values.

In Fig.~\ref{fig:intensity scatter plot}, we showcase the benefit of multispectral information for tree species classification by illustrating scatter plots of the mean intensity of each segment for different combinations of the three channels in the HeliALS and Optech Titan datasets. Importantly, the intensity data at multiple wavelengths significantly helps to distinguish certain pairs of species, which is especially clear for the HeliALS data in Figs.~\ref{fig:intensity scatter plot}(a)-(c). For example, maple and pine share a relatively similar intensity distribution for channel 3, but the intensity of channel 1 efficiently sets apart these species based on Fig.~\ref{fig:intensity scatter plot}(b). On the other hand, birch shows a clearly lower mean intensity for channel 3 compared to any other species in the HeliALS dataset as shown in, e.g., Fig.~\ref{fig:intensity scatter plot}(c). 

\begin{figure*}[ht]
    \centering
    \includegraphics[width=0.85\textwidth]{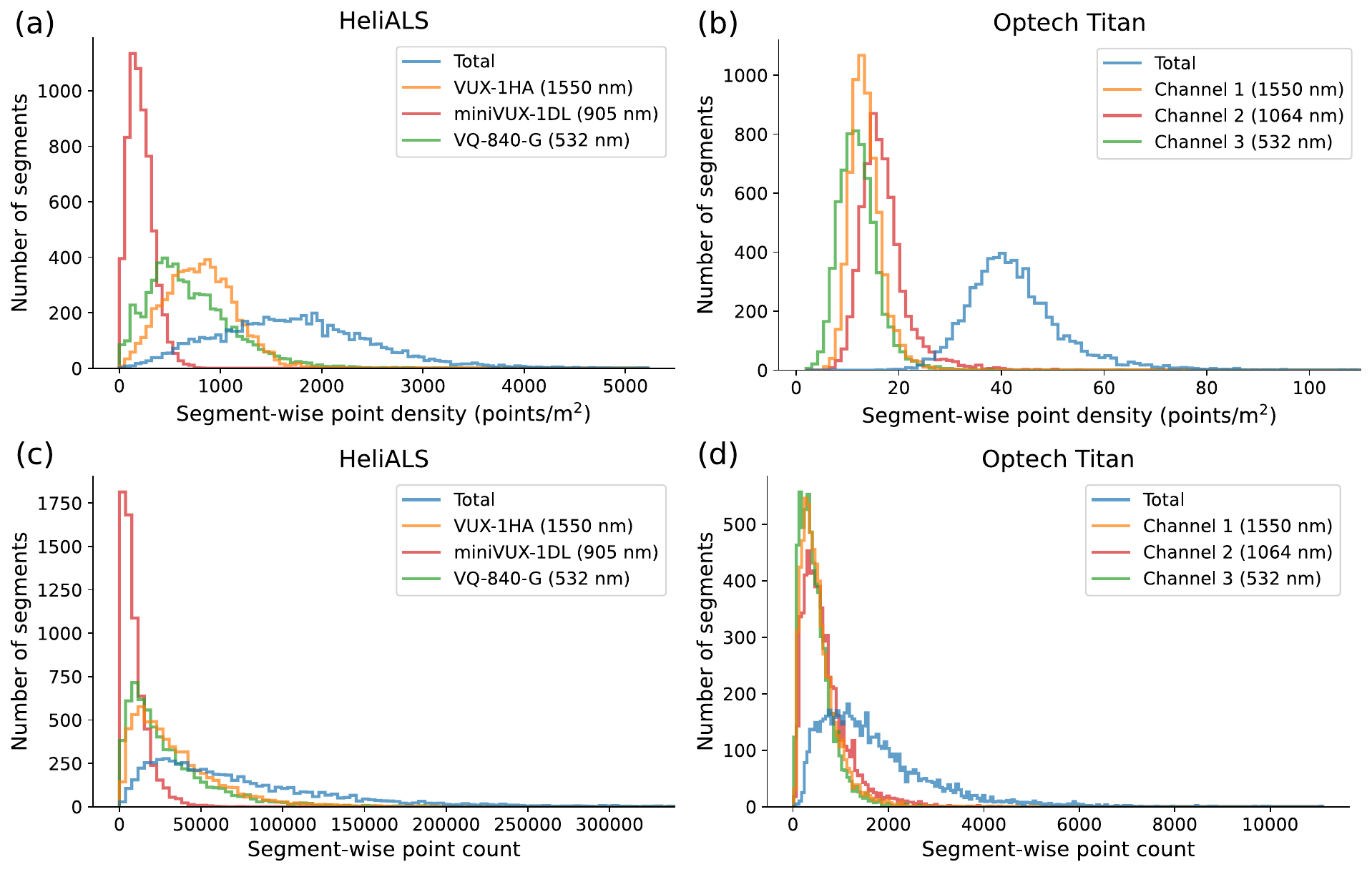}
    \caption{Channel-wise distribution of point count and density within the tree segments of the MS-ALS-SPECIES dataset. (a,b) Histogram of segment-wise point densities in the multispectral (a) HeliALS and (b) Optech Titan datasets. (c, d) Histogram of segment-wise point counts in (c) the HeliALS and (d) Optech Titan datasets. The point density and point count histograms denoted with ``Total'' have been evaluated for point clouds containing points from all scanners and channels in the HeliALS and Optech Titan datasets, respectively.}
    \label{fig:segment-wise point densities and counts}
\end{figure*}

The green wavelength channel in ALS systems is highly susceptible to noise from ambient illumination and atmospheric scattering.
This typically results in a significant volume of noise points midair between the forest canopy and the acquisition platform (e.g., aircraft or helicopter).
To address this, tree segments from both the HeliALS and Optech Titan datasets were filtered for vertical outliers before being saved in LAS format.
First, a subset of points, excluding noise points, was selected from all points located within the boundaries of the watershed segments (defined as polygons in the Easting–Northing plane).
Then, the minimum and maximum $z$-coordinate values were computed for the non-noise points. Finally, a margin of 0.05 meters below and above the minimum and maximum $z$-coordinate values was used to filter out points from the full set of points inside the watershed segment boundaries.

The HeliALS point clouds were collected without a gyro-stabilized sensor platform during a partly cloudy summer day with noticeable convective turbulent activity. The attitude of the data acquisition helicopter varied considerably along the longitudinal axis during the measurement campaign, which affected the strip coverage on the ground surface.
Therefore, some of the tree segments might contain zero points from one or more laser scanners. The scanners had different scanning directions and fields of view as summarized in Table~\ref{tab: HeliALS specs}, and therefore, they were differently impacted by this issue. The scanners that were mostly affected were the miniVUX-1DL and VQ-840-G, due to their Palmer-type scanning geometry. In detail, 6272 tree segments contained points from all three scanners, 33 tree segments were missing points from the VQ-840-G scanner, one tree segment was missing points from the miniVUX-1DL scanner, and 20 tree segments were missing points from both of the aforementioned scanners.
The histograms of segment-wise point counts and densities are visualized in Fig. ~\ref{fig:segment-wise point densities and counts}. 

The segments with few or no points for miniVUX-1DL and VQ-840-G scanners can be dealt with various imputation strategies. Here, we summarize the strategy used by the best performing deep learning method in Sec.~\ref{sec: results benchmarking species classsification methods}, FGI-PointTransformer-DL-3D: If a tree segment contains points from all of the three channels but only a few points for some of these channels, nearest neighbor interpolation of point attributes may suffer from locally missing point attributes for the sparse channels. This can be mitigated by computing the median of the existing values for the locally missing point attributes within the entire tree segment and using the median as the imputation value for the  interpolation. If there are no points for a given channel in the entire tree segment, then the imputation value can be obtained from a set of 1000 randomly selected tree segments in the dataset, which ensures that the imputed value comes from the distribution of the dataset.

\section{Tree species classification}
\label{sec: tree species classification}

We briefly summarize the tree species classification algorithms  compared on the MS-ALS-SPECIES dataset in Sec.~\ref{sec: species classification methods}. Furthermore, Sec.~\ref{sec: accuracy analysis} presents the metrics used to assess the classification accuracy and its confidence intervals, and Sec.~\ref{sec: Further analyses of the classification results}  introduces analyses of the classification results carried out in the present study.

\subsection{Classification algorithms}
\label{sec: species classification methods}

 In this article, we compare the accuracy of tree species classification for the same algorithms as in the benchmarking study conducted on the MS-ALS-SPECIES dataset \citep{taher2025}, but focusing on clean tree segments and the classification accuracy as a function of tree height. The studied species classification methods can be divided into deep learning (DL) approaches and shallow machine learning (ML) classifiers, such as random forest (RF). Deep learning classifiers can be further divided into point-based deep learning (3D DL) approaches and image-based deep learning (2D DL) methods as discussed in Sec.~\ref{sec: related work}. Here, we briefly summarize the implementation of the best performing methods in each category used as examples in Sec.~\ref{sec:results and discussion}. We refer the reader to \citet{taher2025} for a more thorough description of the methods included in the comparison. The classification methods have been trained on the original training set of 1065 tree segments as in \citet{taher2025}.  In general, we follow the  naming convention  \textlangle organization name\textrangle-\textlangle method details\textrangle-\textlangle classifier category\textrangle-\textlangle2D or 3D-based method\textrangle~for the participating methods. As an exception, the variants of the DetailView method do not fit this naming convention as discussed in \citet{taher2025}. All the results in Sec.~\ref{sec:results and discussion} have been obtained by training the classifiers on the training set of the benchmarking study, i.e., using \verb|training-and-test-segments-with-species.csv|.

 On both HeliALS and Optech Titan datasets, FGI-PointTransformer-DL-3D is the best performing 3D DL method. FGI-PointTransformer-DL-3D utilizes the point transformer architecture proposed by \citet{zhao2021point}. On the HeliALS data, the point transformer network of FGI-PointTransformer-DL-3D takes as input a (3+13)D point cloud with 8192 points per segment, where the features include the normalized coordinates, multispectral information in the form of intensities, amplitudes, reflectances, and echo deviations for the three channels, and an echo return number. On the Optech Titan data, the point transformer network takes as input a (3+4)D point cloud with 2048 points per segment, where the features include coordinates, range-corrected intensities for the 3 channels, and an echo return number.

 On the HeliALS dataset, DetailView-DL-2D \citep{Detailview2024, puliti2025benchmarking} is the best performing image-based deep learning model. DetailView-DL-2D was also the best performing model in \citet{puliti2025benchmarking}. DetailView takes as input (4+1+1+1)×2D depth image projections with 256×256 pixels per segment corresponding to four depth images from the sides, a depth image from the top, a depth image from the bottom, and a close-up on the height interval $z\in [1.0, 1.5]$ m. The backbone of DetailView is based on DenseNet-201-instances \citep{huang2017densely}.

FGI-RF-ML represents the only random forest model tested on the HeliALS dataset. In the training phase, FGI-RF-ML computes 92 hand-crafted features for each tree segment, including geometric features based on the channel with most points, and intensity features, such as intensity histograms for each laser channel. The used features are described in more detail in prior works~\citep{lehtomaki2015object, yu2017single, hakula2023individual}.

IBL-Balanced-RF is the best performing classifier on the Optech Titan dataset. For the training phase, tree segments are manually inspected to remove segments affected by under- or over-segmentation, resulting in an updated training set of 660 tree segments. Hand-crafted features are computed following~\citet{kaminska2021single} and \citet{lisiewicz2025comprehensive}, including geometric features based on the full point cloud and intensity features based on individual channels and their combinations.
A balanced random forest technique~\cite{chen2004using} is used to up-sample minority species in the training phase.

 \subsection{Accuracy metrics}
\label{sec: accuracy analysis}

\begin{figure*}[ht]
    \centering
    \includegraphics[width=\textwidth]{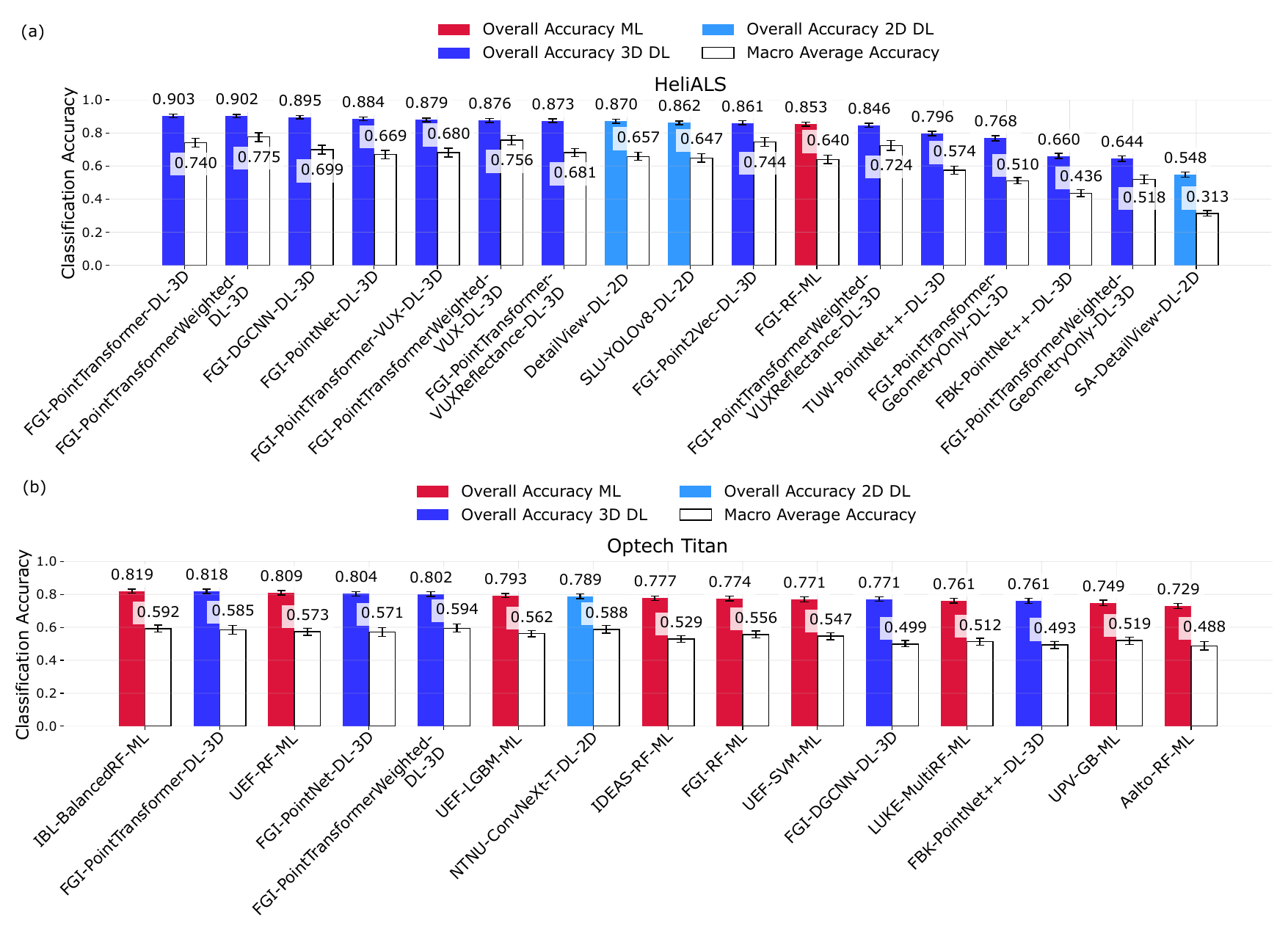}
    \caption{Classification performance of studied species classification algorithms on clean tree segments. (a) Overall accuracy (filled bars) and macro-average accuracy (outlines) of tree species classification across clean tree segments in the test set for methods using the HeliALS dataset in \citet{taher2025}. The method type is indicated by the color of the bar, with point-based DL methods shown in dark blue, image-based DL methods in light blue, and shallow ML classifiers in red. The uncertainty estimates represent 95\% confidence intervals obtained by bootstrapping the test set with replacement.  The model parameters of SA-DetailView-DL-2D were directly transferred from the FOR-species-20K dataset and were thus not trained on the present dataset. (b) Same as (a) but for the methods using the Optech Titan dataset. Names of the participating research organizations corresponding to the abbreviations are listed in the Acknowledgements and in \citet{taher2025}.}
    \label{fig:OA_for_clean_trees}
\end{figure*}

We evaluate the accuracy of the classification methods using overall accuracy and macro-average accuracy as our main evaluation metrics. The overall accuracy (OA) corresponds to the proportion of correctly classified tree segments among all segments
\begin{equation}
    \text{OA} (y, \hat{y}) = \frac{1}{n_{\text{samples}}} \sum_{i = 1}^{n_{\text{samples}}} \mathds{1}(\hat{y}_i = y_i),
\end{equation}
where $\hat{y}_i$ denotes the predicted species of the $i$th segment, $y_i$ is the corresponding ground-truth label, $\mathds{1}(\cdot)$ is an indicator function and $n_{\text{samples}}$ is the number of  tree segments.
In contrast to overall accuracy, macro-average accuracy weighs each species class equally
\begin{equation}
    \text{Macro-average accuracy} = \frac{1}{|S|} \sum_{s \in S} \frac{\text{TP}_s}{\text{TP}_s + \text{FN}_s},
\end{equation}
where $|S|=9$ is the number of tree species in the dataset, and $\text{TP}_s$ and $\text{FN}_s$ denote the number of true positives and false negatives of species $s$, respectively.

Furthermore, we estimate 95\% confidence intervals of the overall accuracy and macro-average accuracy for each of the classification algorithms by bootstrapping the test set with replacement 2000 times. The confidence intervals describe the uncertainty arising from the finite size of the test set, but do not capture potential randomness in the model training. See Sec. 3.3 in \citet{taher2025} for more details on the estimation of confidence intervals. 

\subsection{Further analyses of the classification results}
\label{sec: Further analyses of the classification results}

In the present study, we benchmark the classification methods on clean tree segments, study the impact of tree height on species classification accuracy, and inspect some typical misclassifications. As described in Sec. ~\ref{sec: ground-truth reference data}, clean tree segments correspond to high-quality segments that belong to the profile category ‘Single tree’ and do not belong to the crown category ‘Smaller tree next to larger tree’. We evaluated the overall accuracy and macro-average accuracy for each classification method on the clean segments of the test set. The impact of tree height on classification accuracy was studied by estimating the overall accuracy as a function of tree height. This analysis was carried out for the best performing 3D DL, 2D DL and ML methods. Furthermore, the mean accuracy across all methods was computed. Results based on the HeliALS and Optech Titan datasets were analysed separately. Finally, two common misclassification types, i.e., aspens incorrectly classified as birches and alders incorrectly classified as aspens, were selected for further study, which included visual inspection of tree segments misclassified by several classifiers and re-visiting some trees in the field to better understand the reasons for the misclassifications.

\section{Results and discussion}
\label{sec:results and discussion}

Here, we study the classification accuracy of the methods participating in \citet{taher2025} on clean tree segments in Sec.~\ref{sec: results benchmarking species classsification methods} and as a function of tree height in Sec.~\ref{sec: results vs tree height}. These results act as a follow-up study on \citet{taher2025}, providing further insight into the factors affecting the classification accuracy and the ranking of the methods with minimal segmentation errors.
In Sec.~\ref{sec: inspection of segments with consistent misclassifications}, we further discuss reasons behind systematic misclassification of certain tree segments by multiple different classification methods.

\subsection{Benchmarking classification methods on clean tree segments}
\label{sec: results benchmarking species classsification methods}

\begin{figure*}[ht]
    \centering
    \includegraphics[width=\textwidth]{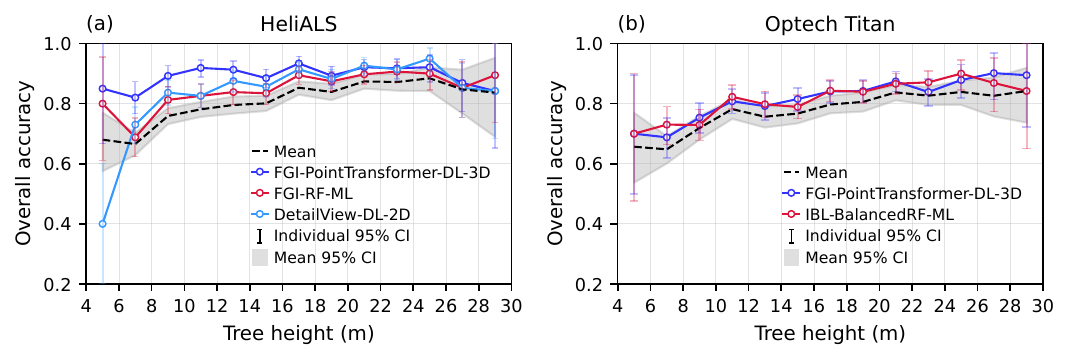}
    \caption{Classification accuracy of the best performing DL and ML methods as a function of tree height. (a) Overall accuracy on the HeliALS dataset calculated across clean tree segments as a function of tree height for the best performing 3D DL (solid dark blue line), 2D DL (solid light blue line), and ML (solid red line) methods together with the mean across all methods (dashed line). Error bars indicate bootstrapped 95\% confidence intervals for the accuracy estimates of individual methods, while the shaded gray region indicates the bootstrapped 95\% confidence interval for the mean accuracy across all methods. (b) Same as (a) but showing the overall accuracy on the Optech Titan dataset for the best performing 3D DL (dark blue) and ML (red) methods together with the mean across all methods. 
    }
    \label{fig:height_accuracy_CI}
\end{figure*}

We first compare the overall and macro-average classification accuracy of the methods participating in \citet{taher2025} for clean tree segments of the test set with nearly ideal segmentation.  This is in contrast to \citet{taher2025} that considered the full test set including some non-ideally segmented trees. Further details on the clean tree segments  and their selection can be found in Fig.~\ref{fig:statistics} and Sec.~\ref{sec: ground-truth reference data}. As shown in Fig.~\ref{fig:OA_for_clean_trees}, the point transformer models, FGI-PointTransformer-DL-3D and FGI-PointTransformerWeighted-DL-3D, top the list on the HeliALS data and the random forest model IBL-BalancedRF-ML ranks first on the Optech Titan dataset, which agrees with the results of \citet{taher2025}. On the HeliALS dataset, FGI-PointTransformer-DL-3D reaches an overall accuracy of 90.3\%, whereas FGI-PointTransformerWeighted-DL-3D achieves a macro-average accuracy of 77.5\%. On the sparser  Optech Titan dataset, IBL-BalancedRF-ML reaches an overall accuracy of 81.9\% and a macro-average accuracy of 59.2\%, which is significantly lower than the best results reached on the denser HeliALS dataset.

On both datasets, a few of the best performing methods are within confidence intervals. For HeliALS data, the median width of the 95\% confidence interval (CI)  was 0.024 for overall accuracy and  0.051 for macro-average accuracy. 
For Optech Titan data, the median 95\% CI width  was 0.028 for overall accuracy and 0.046 for macro-average accuracy.

For most methods, the exclusion of non-ideal tree segments in the test set improved the overall accuracy and macro-average accuracy  by approximately 2--3 percentage points compared to \citet{taher2025}. Generally, there are only minor changes in the ranking of the methods, especially considering the best and worst performing methods. This provides a further verification on the results presented in \citet{taher2025}. The largest change in the ranking is the rise of NTNU-ConvNeXt-T-DL-2D by three positions, though this is mostly explained by several methods achieving nearly identical classification accuracies within the confidence intervals of each other.

\subsection{Impact of tree height on species classification accuracy}
\label{sec: results vs tree height}


\begin{figure*}[ht]
    \centering
    \includegraphics[width=\textwidth]{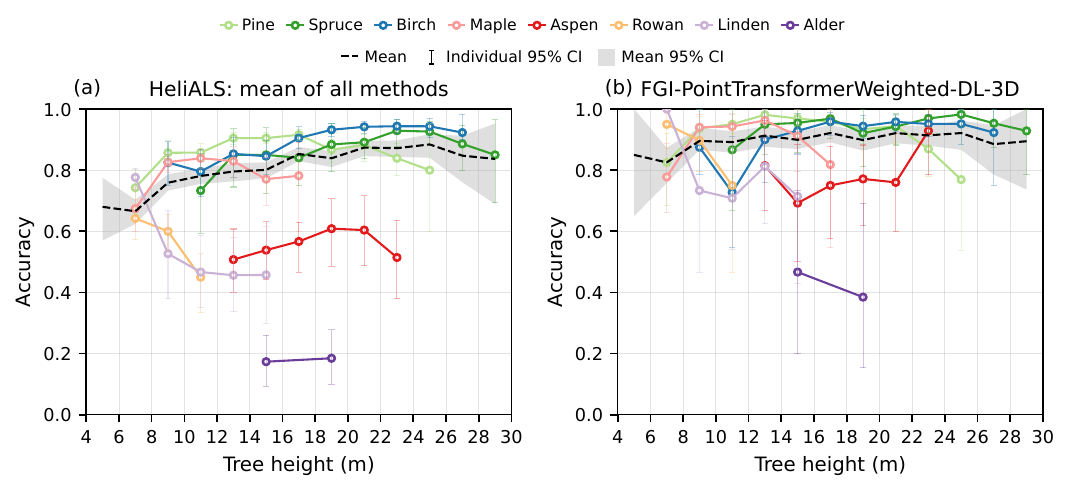}
    \caption{ 
    Height dependence of species-wise classification accuracy on the HeliALS dataset. (a) Overall accuracy per species (solid lines) and for all clean tree segments (dashed line) as a function of tree height averaged across all classification methods using the HeliALS dataset. Only clean tree segments of the test set are considered in the evaluation of the overall accuracy. For a given 2-m height interval, the accuracy is computed for each species with at least 10 samples in the test set. The results for oak are not shown since this criterion is not satisfied for any height interval.  Error bars and shaded regions indicate bootstrapped 95\% confidence intervals estimated by resampling tree-level predictions with replacement within each 2-m height interval.  (b) Same as (a) but for FGI-PointTransformerWeighted-DL-3D, the  method that achieved the highest macro-average accuracy on the HeliALS data.}
    \label{fig:species_HeliALS_1}
\end{figure*}


\begin{figure*}[htb!]
    \centering
    \includegraphics[width=\textwidth]{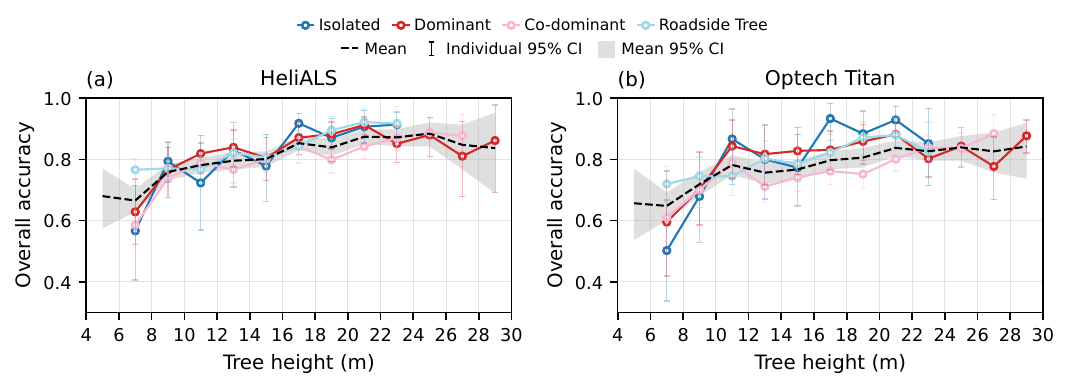}
    \caption{Height dependence of classification accuracy for different crown categories. (a) Overall accuracy for each crown category (solid lines) averaged across all classification methods using the HeliALS dataset as a function of tree height. 
    The mean overall accuracy considering all clean tree segments is shown with the dashed line. Error bars indicate bootstrapped 95\% confidence intervals for the crown-category accuracy estimates, while the shaded region indicates the bootstrapped 95\% confidence interval for the mean overall accuracy curve. (b) Same as (a) but for the Optech Titan dataset. 
    }
    \label{fig:categories_optech_clean}
\end{figure*}

To understand the impact of tree attributes on classification accuracy, we further examine the overall accuracy as a function tree height for the 
best performing methods of each type as shown in Fig.~\ref{fig:height_accuracy_CI}. We also evaluate the mean accuracy across all methods on both HeliALS and Optech Titan datasets, excluding DetailView-DL-2D, and the variants of the point transformer model with different attribute and reflectance combinations (FGI-PointTransformer-VUX-DL-3D, FGI-PointTransformerWeighted-VUX-DL-3D, FGI-PointTransformer-VUXReflectance-DL-3D, FGI-PointTransformerWeighted-VUXReflectance-DL-3D, FGI-PointTransformer-GeometryOnly-DL-3D, and FGI-PointTransformerWeighted-GeometryOnly-DL-3D).  
To estimate the overall accuracy as a function of height, we divide the tree heights into 2-meter bins and only consider height bins with 10 or more observations.

\begin{figure*}[ht]
    \centering
    \includegraphics[width=0.8\textwidth]{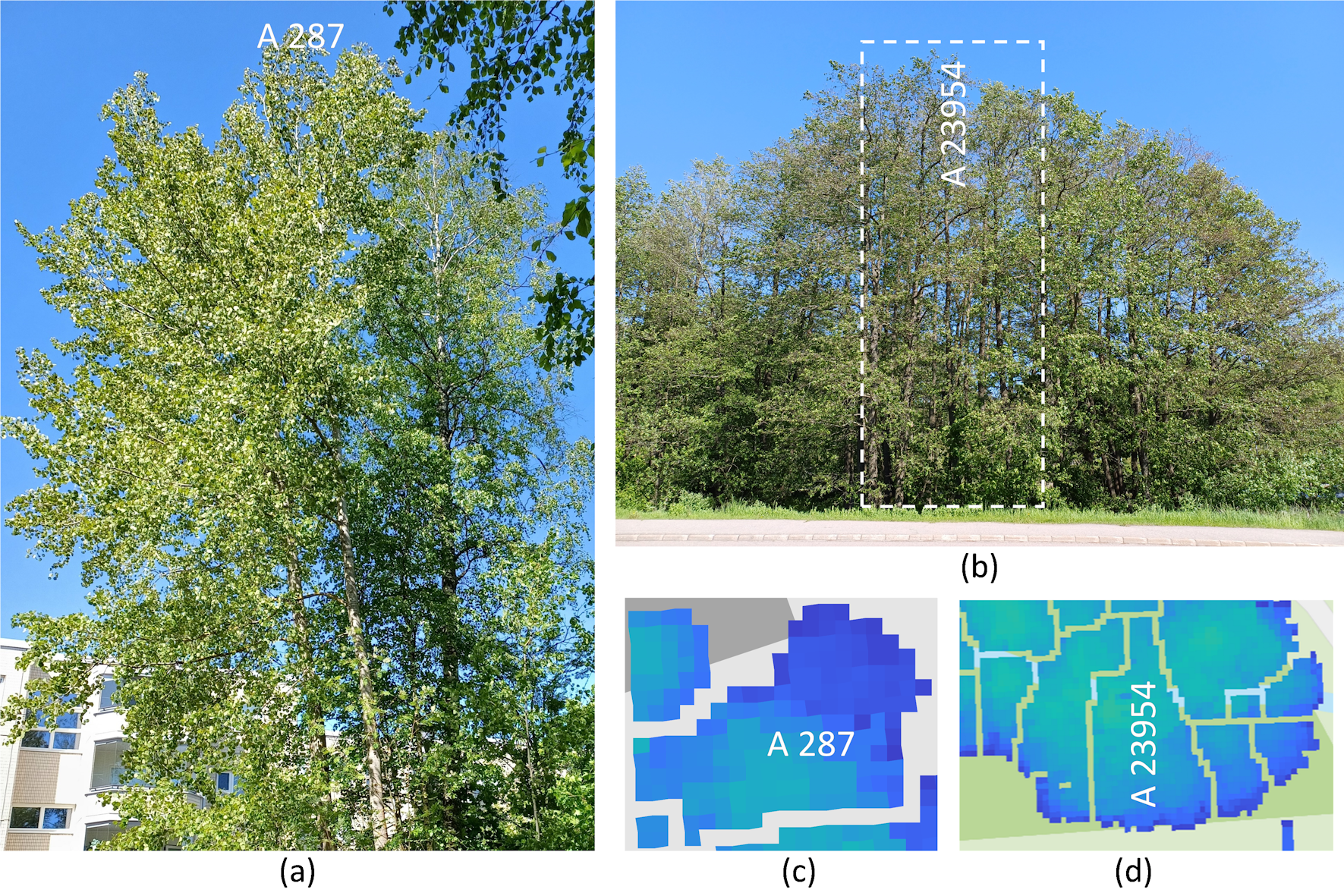}
    \caption{Two examples of segments misclassified by several algorithms. (a) Tree segment A 287 with reference class aspen and misclassified as birch includes several trees, also a high birch, which is in the background behind aspens in the picture. The photograph shows the canopies of the trees. (b) Tree segment A 23954 with reference class alder and misclassified as aspen includes sections of several alders and some lower vegetation with other species. The dashed line shows the approximate location of the segment in the picture. (c) A screen capture from the crowdsourcing tool showing the canopy height model for segment A 287. (d) A screen capture from the crowdsourcing tool showing the canopy height model for segment A 23954 and its neighboring segments visible in panel (b). The background map in the crowdsourcing application in (c) and (d) is Background map (vector) 08/2025 from the National Land Survey of Finland.}
    \label{fig:aspen_alder}
\end{figure*}

Based on Fig.~\ref{fig:height_accuracy_CI}, we observe that the overall accuracy is clearly lower for shorter trees and generally increases with tree height. On the HeliALS data, DetailView-DL-2D representing an image-based DL method and  FGI-RF-ML representing a ML method perform relatively similarly for tree heights exceeding 6~m. In contrast, the best performing point-based DL model, FGI-PointTransformer-DL-3D, is significantly better at predicting the species for small trees. For trees with a height below 15 m, the overall accuracy is up to 10 percentage points higher for the point transformer model than for FGI-RF-ML or DetailView-DL-2D. For trees with a height exceeding 20 m, the three selected methods perform nearly equally. On the Optech Titan dataset, the best DL method, FGI-PointTransformer-DL-3D, and the best ML method, IBL-BalancedRF-ML, behave qualitatively similarly as a function of height. As observed in the scaling analysis of \citet{taher2025}, we expect the deep learning models to increasingly outperform machine learning models on larger training datasets beyond the training set of 1065 tree segments considered here. 

Since the distribution of tree height varies between different tree species as shown in Fig.~\ref{fig:statistics}, we further investigate if the height dependence of classification accuracy is mostly caused by the minority species with a lower classification accuracy being shorter on average. To this end, Figure~\ref{fig:species_HeliALS_1} shows the classification accuracy using the HeliALS data for different species at each 2-meter height bin ranging from 6 to 30 meters. For the HeliALS dataset, we compare the overall accuracy per species averaged across all methods in Fig.~\ref{fig:species_HeliALS_1}(a) and the overall accuracy per species for FGI-PointTransformerWeighted-DL-3D in Fig.~\ref{fig:species_HeliALS_1}(b). By considering the mean across all methods in Fig.~\ref{fig:species_HeliALS_1}(a), we can infer that the height dependence of classification accuracy has two main causes. First, some of the minority species, such as linden and rowan are significantly shorter than the majority species, with their heights rarely exceeding 15 m. Since the overall classification accuracy is lower for these minority species, the overall accuracy considering all species is reduced for shorter trees. 
Second, the classification accuracy of the majority species, such as spruce, pine, and birch,  appears to generally increase with tree height when averaged across all methods.  Thus, the height dependence of classification accuracy appears to be partly caused by the minority species being shorter and partly by an improvement in the classification accuracy of the majority species with increasing height.

Further investigation on the minority tree species with the smallest numbers of trees, i.e., rowan, oak, linden, and alder, shows that the distribution of tree height for rowan, linden and oak is skewed towards short trees as shown in Fig.~\ref{fig:statistics}(b). All of these three tree species have a large proportion of single trees in the dataset, and among other profile categories, ‘Tree section’ is the most typical category, see Fig.~\ref{fig:statistics}(c). Thus, rowan, oak and linden seem to share some similar canopy and height characteristics with each other and with maple. This could also explain the results in Fig.~10(c) of \citet{taher2025}, which showed that the most typical misclassification for rowan, oak and linden segments was with maple. In addition, there were mutual misclassifications between those three minority tree species.

 For the tallest trees approaching a height of 30 m, the classification accuracy slightly decreases, especially for pine trees.  Interestingly, the drop in the overall accuracy of large pines is systematically observed across several methods and datasets as shown in Figs.~\ref{fig:species_HeliALS_1} and ~\ref{fig:species_Optech_1} in Appendix \ref{ap: accuracy vs height}. Based on these figures, the classification accuracy of pines begins to decline for trees taller than 22 meters, with a starker decline starting for pines taller than 24 meters. In total, there are 18 pines taller than 24 meters and they all belong to either dominant or co-dominant crown class.   These tree segments are mostly located in tree clusters or forested areas, which is a possible explanation to the lower classification accuracy. Namely, trees that are more isolated or roadside trees are generally easier to classify due to a cleaner segment and less obstructions as demonstrated by Fig.~\ref{fig:categories_optech_clean}. The profile projection images of the tall pines were examined and no substantial difference to the rest of the segments was detected. Upon observing an orthophoto of the tall pine segments, one of the pines was noticed to have been a misclassified spruce, but this is insufficient to explain the drop in the performance, e.g., for FGI-PointTransformerWeighted-DL-3D.

\subsection{Inspection of segments misclassified by several classifiers}
\label{sec: inspection of segments with consistent misclassifications}

A small proportion of tree segments was consistently misclassified by several of the classification algorithms. Based on the confusion matrices in \citet{taher2025}, we selected two common misclassification types for further inspection: aspens incorrectly classified as birches and alders incorrectly classified as aspens. Birches, aspens, and alders are common species in Finnish forests, and therefore, it is important to understand the reasons behind these classification errors. We focused on 20 aspen segments classified as birches and 9 alder segments classified as aspens. These were misclassified by three well-performing classifiers, namely, FGI-RF-ML trained on HeliALS data, FGI-PointTransformerWeighted-DL-3D trained on HeliALS data, and FGI-PointTransformer-DL-3D trained on Optech Titan data.  
We visually inspected the corresponding tree segments  on the screen using the crowdsourcing tool and rotating point cloud visualizations created from the HeliALS data. We also considered the profile categories and crown categories 
of the segments and comments written during the field work. In addition, we used Google Street View to inspect the trees visible near the roads. Finally, we re-visited and photographed a few of the trees  in the field.

\begin{figure*}[ht]
    \centering
    \includegraphics[width=0.8\textwidth]{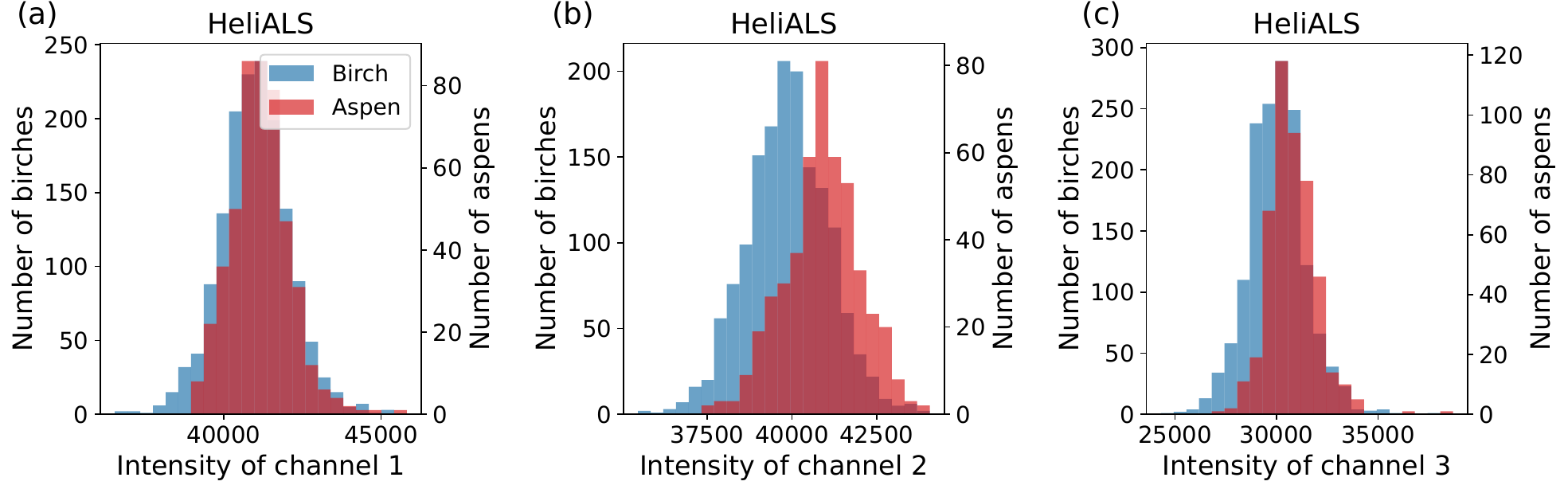}
    \caption{Spectral separability analysis for birch and aspen tree segments in the HeliALS dataset, showing the distribution of the mean intensities of (a) channel 1, (b) channel 2, and (c) channel 3 for each birch and aspen tree segment.}
    \label{fig:birch_vs_aspen_intensity_helials}
\end{figure*}

\begin{figure*}[ht]
    \centering
    \includegraphics[width=0.8\textwidth]{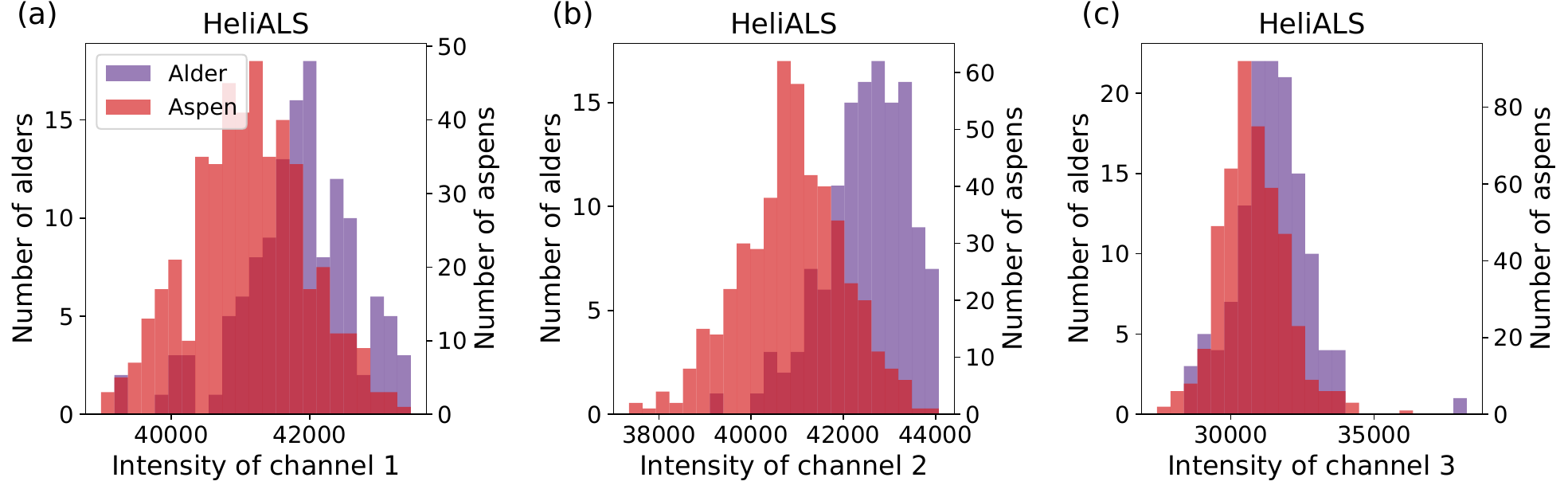}
    \caption{Spectral separability analysis for aspen and alder tree segments in the HeliALS dataset, showing the distribution of the mean intensities of (a) channel 1, (b) channel 2, and (c) channel 3 for each aspen and alder tree segment.}
    \label{fig:alder_vs_aspen_intensity_helials}
\end{figure*}

According to our findings, most aspen segments misclassified as birches were challenging segments to classify. Namely, several trees were often growing within the segments, and some of the segments even included birches in addition to aspens, see Fig.~\ref{fig:aspen_alder}(a). In certain cases, the field surveyor has had  difficulties to determine the reference class for the segment due to a high density of trees or possible changes in the field after the Optech Titan data acquisition used as the basis for the segmentation. It is likely that at least one or two of the misclassified segments had an incorrect reference class. The classification of aspen and birch segments is further hampered by their relatively similar spectral response as shown  in Fig.~\ref{fig:birch_vs_aspen_intensity_helials} for the HeliALS dataset. Figures~\ref{fig:birch_vs_aspen_intensity_helials}(a) and (c) show that the intensity distributions of these species greatly overlap for channels 1 and 3. However, the distributions of the two species differ somewhat for the intensity of channel 2 shown in Fig.~\ref{fig:birch_vs_aspen_intensity_helials}(b), though there is still significant overlap. Similar observations can be made from the spectral separability analysis for the Optech Titan dataset summarized in Appendix \ref{ap: spectral separability analysis Optech Titan}.

Alders in the area tend to grow in dense clusters as shown in Fig.~\ref{fig:aspen_alder}(b), leading to challenges in segmentation and also in labeling of the segments in the field. Most of the alder segments misclassified as aspens contained sections of several alders. In addition to large alders, some of the segments included lower vegetation with other species. Furthermore, alders and aspens also have relatively similar spectral characteristics especially for channels 1 and 3 as shown in Fig.~\ref{fig:alder_vs_aspen_intensity_helials}, which partially explains the misclassification of alder segments as aspen trees. 

Overall, the findings suggest that the consistent misclassification of some aspen segments as birch and some alder segments as aspen can be mainly attributed to the segments being challenging with multiple trees rather than to the inability of the classification methods to identify species of individual trees. The segmentation of individual trees from ALS data in dense forest areas is a challenging task, and thus, improved segmentation methods have potential to contribute to improved accuracy of species classification in fully automatic workflows in the future. In addition to non-ideal segmentation, the similarities in the spectral response for the considered species pairs further complicates the classification.

\section{Conclusions}
\label{sec:conclusions}

This article presents the first open multispectral lidar dataset for tree species classification, aiming to advance research in this field and the development of  deep learning methods on multispectral data. In Sections~\ref{sec:dataset}, \ref{sec: ALS acquisition}, and \ref{sec: open dataset}, we have provided details on the data collection and a user guide on the dataset that was originally used for the international benchmarking study by \citet{taher2025}. The open dataset provides three-wavelength multispectral ALS data at two point densities ($\sim 35$~$\mathrm{pts}/\mathrm{m}^2$ and $\sim 1300$~$\mathrm{pts}/\mathrm{m}^2$) for 6326 tree segments together with ground truth species labels and other properties, such as height, crown class, and profile category describing the segmentation quality. In Sec.~\ref{sec:dataset}, we have described a novel approach for field data collection using a crowdsourcing tool that was developed as a part of the study and applied to collect ground truth species labels in an efficient and scalable manner. 

To provide further insights into the dataset and the behavior of the classification methods, we carried out further analyzes on tree species classification using multispectral ALS data in Sec.~\ref{sec:results and discussion}.  Using a subset of the test set consisting of "clean tree segments", i.e., individual trees with nearly ideal segmentation, we demonstrated that the classification accuracy was improved by 2--3 percentage points compared to the results of \citet{taher2025}. In other words, this represented a 15\% reduction in the overall classification error compared to the full test set. Importantly, the ranking of the classification methods remained largely unchanged and the point transformer model still achieved the highest classification accuracy, thus confirming the results of \citet{taher2025}. 

We further studied the relation of classification accuracy and tree height across all clean tree segments, as well as within each species and crown class. We observed that the classification accuracy generally increases with tree height, and the point transformer model representing a point-based DL model helps to significantly increase the classification accuracy for minority species and short trees.  For trees under 15 m in height, the point transformer model achieved an overall accuracy of up to 10 percentage points higher than a random forest classifier or the image-based DetailView model.  
Finally, we investigated reasons behind the systematic misclassification of a small fraction of tree segments by several classifiers participating in the benchmarking study by \citet{taher2025}. Based on our observations, consistent misclassification  can be mainly attributed to specific challenges with the segments rather than to the inability of the classification methods to identify species of individual trees. 

\section*{Data availability}

MS-ALS-SPECIES dataset is publicly available in Zenodo: \url{https://doi.org/10.5281/zenodo.17077255}. The dataset includes both HeliALS and Optech Titan point clouds for the studied tree segments together with ground-truth species labels.

\section*{Acknowledgements}

We gratefully acknowledge the Research Council of Finland (RCF), the NextGenerationEU instrument and Finnish Government (VA-MMM) funding for the following projects ``Collecting accurate individual tree information for harvester operation decision making'' (RCF 359554), “High-performance computing allowing high-accuracy country-level individual tree carbon sink and biodiversity mapping” (RCF 359203), ``Forest-Human-Machine Interplay -- Building Resilience, Redefining Value Networks and Enabling Meaningful Experiences'' (RCF 359175), “Measuring Spatiotemporal Changes in Forest Ecosystem” (decision number 346382)'', ``Digital technologies, risk management solutions and tools for mitigating forest disturbances'' (RCF 353264, NextGenerationEU), and the Ministry of Agriculture and Forestry grant ``Future Forest Information System at Individual Tree Level 2.0'' (VA-MMM-2024-25-1). 

We also gratefully acknowledge the following partners of the benchmarking study, who participated in the results towards Fig.~\ref{fig:OA_for_clean_trees}: Norwegian University of Science and Technology (NTNU), Aalto University (Aalto), University of Eastern Finland (UEF), Norwegian Institute for Bioeconomy Research (NIBIO), Natural Resources Institute Finland (LUKE), IDEAS NCBR, Forest Research Institute (IBL), Swedish University of Agricultural Sciences (SLU), Universitat Politècnica de València (UPV), Technische Universität Wien (TUW), and Bruno Kessler Foundation (FBK). Furthermore, we acknowledge the help of other FGI personnel and University of Helsinki partners for participating in the collection of the field reference dataset using the crowdsourcing application. Individual persons participating in the benchmarking study or the field data collection can be found from the author list of \citet{taher2025}.

\section*{Author contributions}

M. Hyyppä developed the browser-based crowdsourcing application for field reference collection. M. Hyyppä and K. Salolahti jointly analyzed the benchmarking results, classified tree segments into the different categories, and prepared most of the figures. E. Hyyppä planned the content of the manuscript, carried out the literature review, and led the analysis work and the write-up of the paper. J. Taher developed the FGI deep learning methods for both datasets and curated the dataset for sharing in Zenodo. X. Yu performed the segmentation with the Optech data, and developed the FGI machine learning method for both datasets with help from M. Lehtomäki. L. Matikainen and P. Litkey preprocessed the Optech Titan data and provided ideas for developing the crowdsourcing tool and collecting the reference data. L. Matikainen had an active role in quality checking of the crowdsourced data and participated in the analysis of the results. A. Kukko, H. Kaartinen, and T. Hakala contributed to the data collection using HeliALS and preprocessed the HeliALS data with help from J. Taher and M. Lehtomäki. All authors collected field reference using the crowdsourcing application, supported by University of Helsinki partners and other FGI personnel. E. Hyyppä, M. Hyyppä, J. Taher, K. Salolahti, L. Matikainen, M. Lehtomäki, X. Yu, J. Hyyppä and A. Kukko wrote the manuscript with support from other authors. A. Kukko and J. Hyyppä supervised the project and acquired funding.

\clearpage
\normalsize
\begin{appendices}

\section{List of studies included in Figure~\ref{fig: previous literature}}
\label{ap: list of previous studies}

In this section, we summarize previous ALS-based tree species classification studies included in the analysis of Fig.~\ref{fig: previous literature}. We list previous studies  using single-channel ALS in Table~\ref{tab: past literature single channel}, previous studies using  multispectral ALS in Table~\ref{tab: past literature multispectral}, and previous studies using ALS with passive multispectral or hyperspectral imaging in Table~\ref{tab: past literature passive hyperspectral}. For each study, we report the country and type of the forest under study, number of trees used to train the classifier, the number of species considered in the study, average point density, classification method, and the achieved overall classification accuracy. Based on the information reported in the papers, we have attempted to interpret the number of trees used to train the classifier as the training set size while excluding the number of trees in the test set.

\begin{table*}[ht!]
\centering
\caption{\label{tab: past literature single channel} List of selected previous studies on tree species classification using single-channel ALS or ULS. As the number of trees, we report the number of trees used to train the classifier. In the classifier column, SVM stands for support vector machine, LDA and QDA denote linear and quadratic discriminant analysis, k-MSN stands for k most similar neighbor, and PCTSCN denotes Point Cloud Tree Species Classification Network.}
\resizebox{\textwidth}{!}{
\begin{tabularx}{1.5\textwidth} {
>{\hsize=1.5\hsize\linewidth=\hsize}X
>{\hsize=0.9\hsize\linewidth=\hsize}X
>{\hsize=0.9\hsize\linewidth=\hsize}X
>{\hsize=0.9\hsize\linewidth=\hsize}X
>{\hsize=\hsize\linewidth=\hsize}X
>{\hsize=0.9\hsize\linewidth=\hsize}X
>{\hsize=0.9\hsize\linewidth=\hsize}X
>{\hsize=\hsize\linewidth=\hsize}X
  }
 \specialrule{1.0pt}{0em}{0em}
  \textbf{Reference} & \textbf{Country} & \textbf{Forest type} & \textbf{Number of trees} & \textbf{Number of species} & \textbf{Point density} $(\mathrm{pts}/\mathrm{m}^2$)&  \textbf{Classifier}  & \textbf{Overall accuracy} ($\%$) \\
 \specialrule{1.0pt}{0em}{0em}
\citet{orka2010effects} & Norway & Boreal forest & 241--390 & 2 & 5 & RF & 86.9--97.1
 \\
\specialrule{1.0pt}{0em}{0em}
\citet{kim2009tree} & United States & Urban forest & 222 & 2 & $<5$--25 & LDA & 73.1--90.6 
 \\
\specialrule{1.0pt}{0em}{0em}
\citet{zhang2013support} & Australia & Cool temperate rainforest & 594 & 2 & 4 & SVM & 92.8
 \\
\specialrule{1.0pt}{0em}{0em}
\citet{korpela2010tree} & Finland & Boreal forest & 10337 & 3 & 7--14 & RF & 83.5--90.8
 \\
\specialrule{1.0pt}{0em}{0em}
\citet{hovi2016lidar} & Finland & Boreal forest & 9930 & 3 & 10 & QDA & 91
 \\
\specialrule{1.0pt}{0em}{0em}
\citet{kukkonen2019multispectral} & Finland & Boreal Forest & - & 3 & 4.8 & LDA & 85 
 \\
\specialrule{1.0pt}{0em}{0em}
\citet{holmgren2008species} & Sweden & Boreal forest & 1711 & 3 & 50 & QDA & 88
 \\
\specialrule{1.0pt}{0em}{0em}
\citet{vauhkonen2010imputation} & Finland & Boreal forest & 1898 & 3 & 7 & k-MSN & 78.9
 \\
\specialrule{1.0pt}{0em}{0em}

\citet{orka2013characterizing} & Norway & Boreal forest & 1089 & 3 & 7.2 & Balanced RF & 74.4
 \\
\specialrule{1.0pt}{0em}{0em} 

\citet{lin2016comprehensive} & Finland & Park forest & 40 & 4 & 10 & SVM & 92.5
 \\
\specialrule{1.0pt}{0em}{0em} 
\citet{amiri2019classification} & Germany & Temperate forest & 269 & 4 & 60 & Logistic regression & 74.5
 \\
\specialrule{1.0pt}{0em}{0em}

\citet{suratno2009tree} & United States & Montane forest & 225 & 4 & 0.44 & LDA & 68
 \\
\specialrule{1.0pt}{0em}{0em}

\citet{you2020forest} & China & Planted forest & 353 & 5 & - & RF & 92.4
 \\
\specialrule{1.0pt}{0em}{0em}

\citet{shi2020improving} & Germany & Temperate forest & 270 & 5 & 70 & RF & 69.3
 \\
\specialrule{1.0pt}{0em}{0em}

\citet{shi2018tree} & Germany & Temperate forest & 215 & 5 & 70 & RF & 65.1
 \\
\specialrule{1.0pt}{0em}{0em}

\citet{liu2017mapping} & Canada & Urban trees & 553 & 15 & 25 & RF & 61.0
 \\
\specialrule{1.0pt}{0em}{0em}

\citet{liu2021tree} & China & - & 960 & 2 & - & LayerNet & 88.8
 \\
\specialrule{1.0pt}{0em}{0em}

\citet{lv2021convex} & China & Temperate forest & 890 & 4 & 40 & PointNet++ & 86.6
 \\
\specialrule{1.0pt}{0em}{0em}

\citet{chen2021classification} & China & Planted temperate forest & 800 & 2 & 270 &  PCTSCN & 92
 \\
\specialrule{1.0pt}{0em}{0em}

\citet{marinelli2022approach} & Italy & Mountainous temperate forest & 906 & 7 & - & multiview CNN & 83.2
 \\
\specialrule{1.0pt}{0em}{0em}

\citet{fan2023tree} & China & - & 438 & 11 & - & PointNet++ & 91.8
 \\
\specialrule{1.0pt}{0em}{0em}

\citet{hakula2023individual} & Finland & Boreal forest & 1473 & 3 & 1400 & RF & 86.6
 \\
\specialrule{1.0pt}{0em}{0em}

\citet{lin2024pctrees} & Kenya & Tropical Savanna & 4000 & 6 & - & Point transformer  & 72
 \\
\specialrule{1.0pt}{0em}{0em}

\citet{puliti2025benchmarking} & Global & - & 17707 & 33 & - & DetailView & 76.9
 \\
\specialrule{1.0pt}{0em}{0em}

\citet{holmgren2004identifying} & Sweden  & Boreal forest & 537 & 2 & - & LDA & 95
 \\
\specialrule{1.0pt}{0em}{0em}

\citet{heinzel2011exploring}  & Germany & Temperate forest & - & 2--6 & 16 & LDA & 91.7 (2 classes), 59 (6 classes)
 \\
\specialrule{1.0pt}{0em}{0em}

\citet{harikumar2020crown} & Italy & Mountainous temperate forest & 270 & 6 & 100 & SVM & 76.6
 \\
\specialrule{1.0pt}{0em}{0em}

\citet{zhong2022identification} & China & Temperate forest & 528 & 6 & 180 & SVM & 76.4
 \\
\specialrule{1.0pt}{0em}{0em}

\citet{quan2023tree} & China & Temperate forest & 814 & 11 & 340 & RF & 60
 \\
\specialrule{1.0pt}{0em}{0em}

\end{tabularx}}
\end{table*}

\begin{table*}[ht!]
\centering
\caption{\label{tab: past literature multispectral} List of selected previous studies on tree species classification using multispectral ALS or ULS. As the number of trees, we report the number of trees used to train the classifier. }
\resizebox{\textwidth}{!}{
\begin{tabularx}{1.5\textwidth} {
>{\hsize=1.5\hsize\linewidth=\hsize}X
>{\hsize=0.9\hsize\linewidth=\hsize}X
>{\hsize=0.9\hsize\linewidth=\hsize}X
>{\hsize=0.9\hsize\linewidth=\hsize}X
>{\hsize=\hsize\linewidth=\hsize}X
>{\hsize=0.9\hsize\linewidth=\hsize}X
>{\hsize=0.9\hsize\linewidth=\hsize}X
>{\hsize=\hsize\linewidth=\hsize}X
  }
 \specialrule{1.0pt}{0em}{0em}
  \textbf{Reference} & \textbf{Country} & \textbf{Forest type} & \textbf{Number of trees} & \textbf{Number of species} & \textbf{Point density} $(\mathrm{pts}/\mathrm{m}^2$)&  \textbf{Classifier}  & \textbf{Overall accuracy} ($\%$) \\
 \specialrule{1.0pt}{0em}{0em}
\citet{kukkonen2019multispectral} & Finland & Boreal Forest & - & 3 & 13.3 & LDA & 88.2 
\\
\specialrule{1.0pt}{0em}{0em}
\citet{yu2017single} & Finland & Boreal Forest & 1167 & 3 & 63 & RF & 85.9
\\
\specialrule{1.0pt}{0em}{0em}
\citet{amiri2019classification} &  Germany & Temperate forest & 269 & 4 & 200 & Logistic regression & 82.1
\\
\specialrule{1.0pt}{0em}{0em}
\citet{axelsson2018exploring} & Sweden & Boreal forest  & 179 & 9 & 30 & LDA & 76.5
\\
\specialrule{1.0pt}{0em}{0em}
\citet{hakula2023individual} & Finland & Boreal forest & 1473 & 3 & 3500 & RF & 90.8
\\
\specialrule{1.0pt}{0em}{0em}
\citet{rana2022effect} & Canada & Boreal forest & 890 & 9 & 15 & RF & 70
\\
\specialrule{1.0pt}{0em}{0em}
\citet{prieur2021comparison} & Canada & Boreal forest & 1180 & 2--12 & 30 & RF & 90.4 (2 classes), 78.6 (4 classes), 53.2 (12 classes)
\\
\specialrule{1.0pt}{0em}{0em}
\citet{wang2024individual} & Canada & Boreal forest & 785 & 6 & 9 & Cross-branch transformer & 83.1
\\
\specialrule{1.0pt}{0em}{0em}
\citet{taher2025} & Finland & Semi-urban boreal forest & 5000 & 9 & 35 and 1300 & Point transformer & 92.0\% (1300 $\mathrm{pts}/\mathrm{m}^2$), 83.7\% (35 $\mathrm{pts}/\mathrm{m}^2$)
\\
\specialrule{1.0pt}{0em}{0em}

\end{tabularx}}
\end{table*}

\begin{table*}[ht!]
\centering
\caption{\label{tab: past literature passive hyperspectral} List of selected previous studies on tree species classification using ALS with passive multispectral or hyperspectral imaging. As the number of trees, we report the number of trees used to train the classifier. }
\resizebox{\textwidth}{!}{
\begin{tabularx}{1.5\textwidth} {
>{\hsize=1.5\hsize\linewidth=\hsize}X
>{\hsize=0.9\hsize\linewidth=\hsize}X
>{\hsize=0.9\hsize\linewidth=\hsize}X
>{\hsize=0.9\hsize\linewidth=\hsize}X
>{\hsize=\hsize\linewidth=\hsize}X
>{\hsize=0.9\hsize\linewidth=\hsize}X
>{\hsize=0.9\hsize\linewidth=\hsize}X
>{\hsize=\hsize\linewidth=\hsize}X
  }
 \specialrule{1.0pt}{0em}{0em}
  \textbf{Reference} & \textbf{Country} & \textbf{Forest type} & \textbf{Number of trees} & \textbf{Number of species} & \textbf{Point density} $(\mathrm{pts}/\mathrm{m}^2$)&  \textbf{Classifier}  & \textbf{Overall accuracy} ($\%$) \\
 \specialrule{1.0pt}{0em}{0em}

\citet{orka2013characterizing} & Norway & Boreal forest & 1089 & 3 & 7.2 & Balanced RF & 87.0
\\
\specialrule{1.0pt}{0em}{0em}

\citet{shi2020improving} & Germany & Temperate forest & 270 & 5 & 70 & RF & 77.4
\\
\specialrule{1.0pt}{0em}{0em}

\citet{shi2018tree}  & Germany & Temperate forest & 215 & 5 & 70 & RF & 76.7
\\
\specialrule{1.0pt}{0em}{0em}

\citet{liu2017mapping} & Canada & Urban trees & 553 & 15 & 25 & RF & 70
\\
\specialrule{1.0pt}{0em}{0em}

\citet{dalponte2014tree} & Norway & Boreal forest & 1002 & 3 & 7 & SVM & 93.5
\\
\specialrule{1.0pt}{0em}{0em}

\citet{deng2016comparison} & Japan & Temperate  forest & 1707 & 4 & 15 & Quadratic SVM & 90.8
\\
\specialrule{1.0pt}{0em}{0em}

\citet{kaminska2018species} & Poland & Temperate forest & 1214 & 6 & 6 & RF & 94.3
\\
\specialrule{1.0pt}{0em}{0em}

\citet{kaminska2021single} & Poland & Temperate forest & 1230 & 8 & 11 & RF & 83
\\
\specialrule{1.0pt}{0em}{0em}

\citet{li2022classification} & China & Temperate forest & 232 & 4 & 10 & SVM & 85.2
\\
\specialrule{1.0pt}{0em}{0em}

\citet{zhong2022identification} & China & Temperate forest & 528 & 5 & 180 & SVM & 89.2 
\\
\specialrule{1.0pt}{0em}{0em}

\citet{quan2023tree}  & China & Temperate forest & 814 & 11 & 340 & RF & 75.7
\\
\specialrule{1.0pt}{0em}{0em}

\citet{shen2017tree} & China & Subtropical forest & 352 & 5 & 10 & RF & 91
\\
\specialrule{1.0pt}{0em}{0em}

\citet{hartling2021urban} & United States & Park forest & 96 & 7 & 900 & RF & 83.3
\\
\specialrule{1.0pt}{0em}{0em}

\citet{dalponte2012tree} & Italy & Mountainous temperate forest & 2156 & 8 & 8.6 & SVM & 83
\\
\specialrule{1.0pt}{0em}{0em}

\citet{hell2022classification} & Germany & Temperate forest & 1380 & 4 & 80 & PointCNN & 87
\\
\specialrule{1.0pt}{0em}{0em}

\citet{mayra2021tree} & Finland & Boreal forest & 2291 & 10.2 & 4 & 3d CNN & 87
\\
\specialrule{1.0pt}{0em}{0em}

\end{tabularx}}
\end{table*}

\clearpage
\section{Further results on classification accuracy as a function of tree height}
\label{ap: accuracy vs height}

\begin{figure*}[ht]
    \centering
    \includegraphics[width=\textwidth]{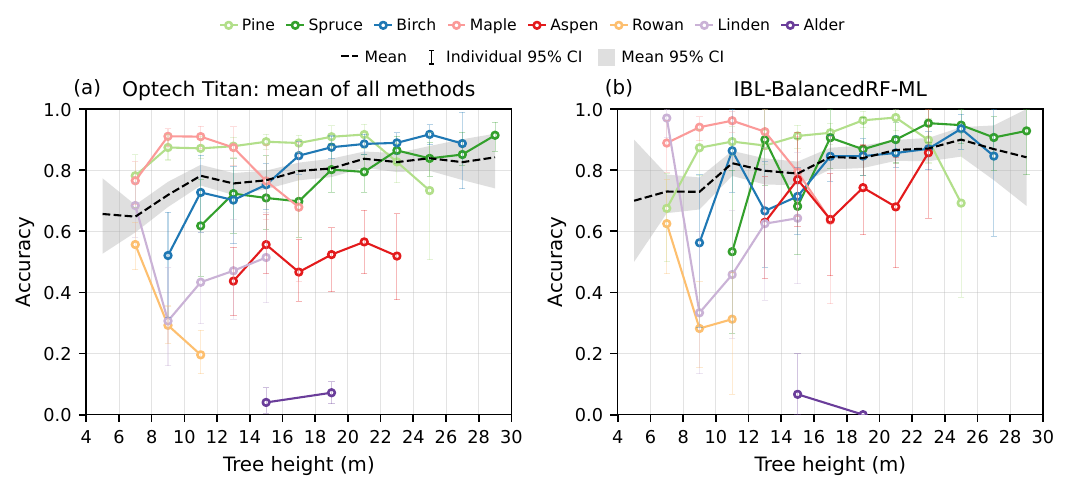}
    \caption{Height dependence of species-wise classification accuracy on the Optech Titan dataset. (a) Overall accuracy per species (solid lines) and for all clean tree segments (dashed line) as a function of tree height averaged across all classification methods using the Optech Titan dataset. Only clean tree segments of the test set are considered in the evaluation of the overall accuracy. For a given 2-m height interval, the accuracy is computed for each species with at least 10 samples in the test set. The results for oak are not shown since this criterion is not satisfied for any height interval.  Error bars and shaded regions indicate bootstrapped 95\% confidence intervals estimated by resampling tree-level predictions with replacement within each 2-m height interval. (b) Same as (a) but for IBL-BalancedRF-ML, the  method that achieved the highest overall accuracy on the Optech Titan data.
    }
    \label{fig:species_Optech_1}
\end{figure*}

Figure \ref{fig:species_Optech_1} shows the classification accuracy using the Optech Titan dataset for different species at each 2 meter height bracket ranging from 6 to 30 meters. The figure shows the accuracy per species as a function of height when averaged across all methods in Fig.~\ref{fig:species_Optech_1}(a) and for the best method in Fig.~\ref{fig:species_Optech_1}(b). The figures only include clean tree segments, and only show the accuracy for height brackets that contain more than 10 trees. 
On the Optech Titan dataset, the differences in overall accuracy between the mean of all methods and the best method are generally smaller compared to the corresponding difference on the HeliALS data visualized in Fig.~\ref{fig:species_HeliALS_1}. The biggest difference in accuracy between the best method and the average are seen for aspens on the Optech Titan dataset. 

\section{Spectral separability analysis of birch, aspen and alder on the Optech Titan dataset}
\label{ap: spectral separability analysis Optech Titan}

\begin{figure*}[ht]
    \centering
    \includegraphics[width=0.8\textwidth]{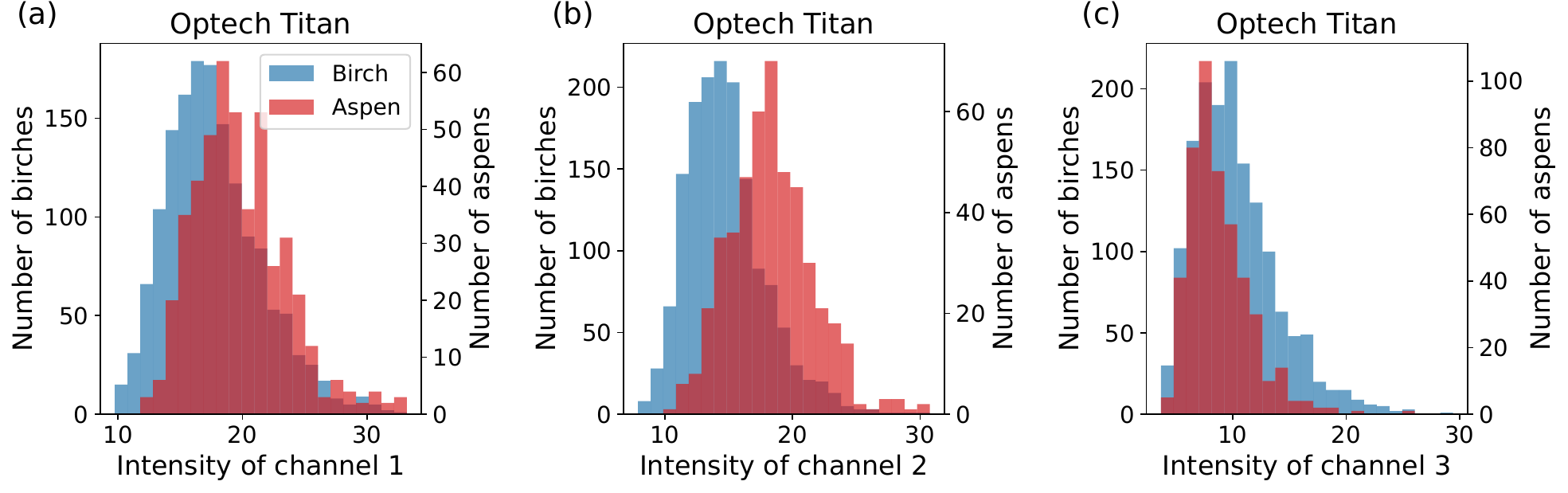}
    \caption{Spectral separability analysis for birch and aspen tree segments in the Optech Titan dataset, showing the distribution of the mean intensities of (a) channel 1, (b) channel 2, and (c) channel 3 for each birch and aspen tree segment.}
    \label{fig:birch_vs_aspen_intensity_titan}
\end{figure*}

\begin{figure*}[ht]
    \centering
    \includegraphics[width=0.8\textwidth]{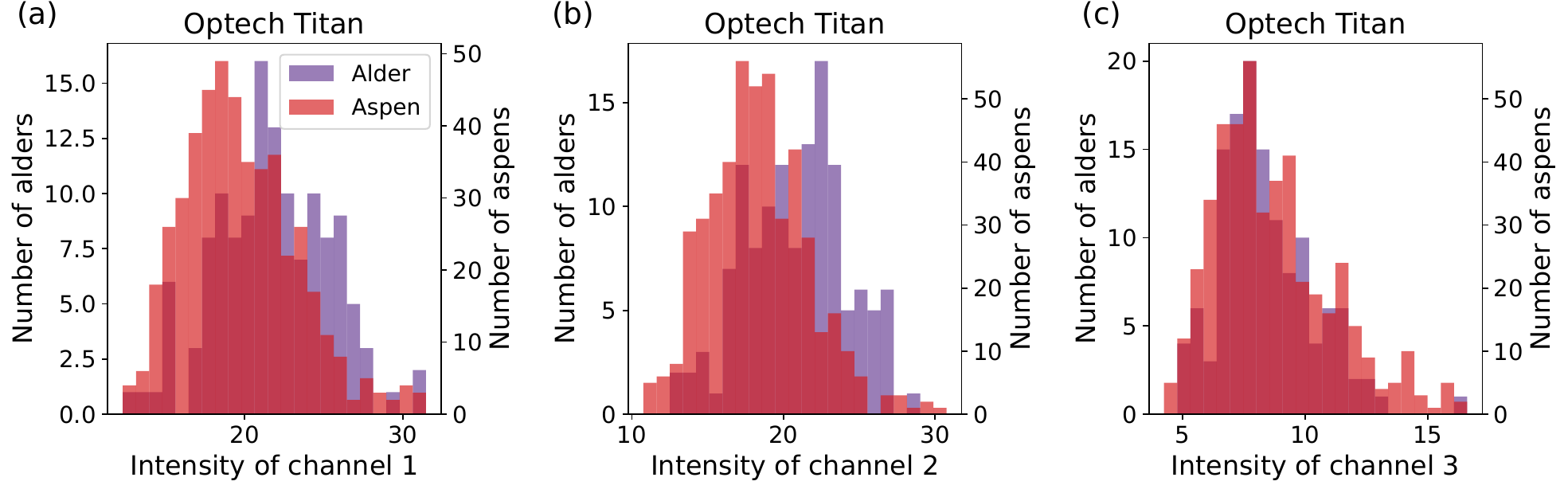}
    \caption{Spectral separability analysis for alder and aspen tree segments in the Optech Titan dataset, showing the distribution of the mean intensities of (a) channel 1, (b) channel 2, and (c) channel 3 for each alder and aspen tree segment.}
    \label{fig:alder_vs_aspen_intensity_titan}
\end{figure*}

Figures~\ref{fig:birch_vs_aspen_intensity_titan} and \ref{fig:alder_vs_aspen_intensity_titan} provide spectral separability analysis for birch, aspen, and alder for the Optech Titan dataset. On our dataset, aspens are most commonly misclassified as birches, while alders are often misclassified as aspens. See the related discussion in Sec.~\ref{sec: inspection of segments with consistent misclassifications} and also Fig. 7 of \citet{taher2025}. To gain further insight into these misclassifications, we compare the spectral response of these tree species across the three channels of the Optech Titan dataset.  Similarly to the spectral separability analyses on the HeliALS dataset in Figs.~\ref{fig:birch_vs_aspen_intensity_helials} and \ref{fig:alder_vs_aspen_intensity_helials}, we observe that there is significant overlap in the spectral characteristics of the three species, especially for channels 1 and 3. Similarly to the HeliALS dataset, the species are best discriminated by the intensity of channel 2, which corresponds to $\lambda \approx 1000$ nm for both sensor systems. However, there is still significant overlap between the intensity distributions, which partially explains why these species are misclassified with each other.

\clearpage

\end{appendices}

{
\footnotesize
\bibliography{references}
}

\end{document}